\documentclass{article}

\usepackage{inputenc}
\usepackage{csquotes}
\usepackage{graphicx}
\usepackage{amsmath} 
\usepackage{tabularray}
\usepackage{tikz}
\usepackage{comment}
\usepackage{pgfplots}[compat=1.17]
\usetikzlibrary{pgfplots.groupplots}
\pgfplotsset{compat=1.18}
\usepackage{xcolor}

\DeclareMathOperator*{\argmin}{arg\,min}
\usepackage{caption}
 \usepackage{float}
 \usepackage{hyperref}

\usepackage[preprint]{neurips_2025}

\usepackage[T1]{fontenc}    %
\usepackage{hyperref}       %
\usepackage{url}            %
\usepackage{booktabs}       %
\usepackage{amsfonts}       %
\usepackage{nicefrac}       %
\usepackage{microtype}      %
\usepackage{xcolor}         %

\title{Do you see what I see? An Ambiguous Optical Illusion Dataset exposing limitations of Explainable AI}

\author{%
Carina Newen 
TU Dortmund University , \\
Research Center Trustworthy Data Science and Security \\
\texttt{carina.newen@cs.tu-dortmund.de} 
 \AND
 Luca Hinkamp \\
 TU Dortmund University, \\
 Research Center Trustworthy Data Science and Security 
 \AND
 Maria Ntonti \\
 TU Dortmund University \\
 \AND
 Emmanuel Müller \\
 TU Dortmund University, \\
 Research Center Trustworthy Data Science and Security \\
 emmanuel.mueller@cs.tu-dortmund.de
}

\begin{document}

\maketitle

\begin{abstract} From uncertainty quantification to real-world object detection, we recognize the importance of machine learning algorithms, particularly in safety-critical domains such as autonomous driving or medical diagnostics. In machine learning, ambiguous data plays an important role in various machine learning domains. Optical illusions present a compelling area of study in this context, as they offer insight into the limitations of both human and machine perception. Despite this relevance, optical illusion datasets remain scarce. In this work, we introduce a novel dataset of optical illusions featuring intermingled animal pairs designed to evoke perceptual ambiguity. We identify generalizable visual concepts, particularly gaze direction and eye cues, as subtle yet impactful features that significantly influence model accuracy. By confronting models with perceptual ambiguity, our findings underscore the importance of concepts in visual learning and provide a foundation for studying bias and alignment between human and machine vision. To make this dataset useful for general purposes, we generate optical illusions systematically with different concepts discussed in our bias mitigation section. The dataset is accessible in Kaggle via \href{https://kaggle.com/datasets/693bf7c6dd2cb45c8a863f9177350c8f9849a9508e9d50526e2ffcc5559a8333}{Ambivision}. Our source code can be found at \url{https://github.com/KDD-OpenSource/Ambivision.git}.
\end{abstract}
\section{Introduction}
\label{sec:intro}
\begin{figure}[h]
    \centering
    \includegraphics[width=6cm]{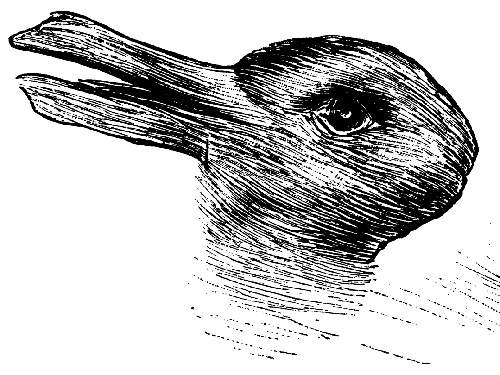}
    \caption{In this image, you can see both a rabbit and a duck. Common XAI methods that highlight important pixels could output exactly the same explanation for either of those classes without improving human understanding of which class was chosen why. This is a critical research gap in explanations- pixel highlighting is simply not enough. One way of distinguishing the two depends on the way you consider the eyes to be looking in. While this is not the only way to approach the problem, we will highlight the usefulness of the gaze for the classification task in our evaluations. However, we show that with one very small addition, we can erase the ambiguity of an image for humans and improve it for machine learners. We argue that the future in XAI lies in uncovering such concepts rather than highlighting pixels, which is the critical research gap we address in this paper.}
    \label{fig:rabbitduck}
\end{figure}

The motivation of this particular optical illusion dataset stems from a novel problem in the explainability domain (XAI). When we discuss explainable AI, uncovering the black-box nature of models, and visualizing the internal workings of a machine learner for image data, research has so far focused on highlighting pixels that are most important or influential in the decision-making process \cite{lime, selvaraju2017grad}. Other approaches highlight important regions \cite{ribeiro2018anchors} or target the most essential features \cite{lundberg2017unified, sundararajan2017axiomatic}. However, these methods often fall short when confronted with perceptual ambiguity- situations where even human interpretation is uncertain. Take a look at Figure \ref{fig:rabbitduck}.
 Depending on the viewer's perception of the eye's direction, the image may be interpreted as either a rabbit or a duck. If you were to use any current XAI algorithm to generate explanations of why it is a rabbit or a duck, the explanations could look exactly the same. They would keep the actual reason behind this optical illusion secret. This is due to the fact that current XAI methods highlight important pixels or areas, which, in this case, are shared by both classes. This suggests that standard XAI methods are currently inadequate when semantic interpretation relies on abstract perceptual cues rather than pixel-level interpretation. Examples of this phenomenon on the provided dataset can be seen in Section \ref{sec:limitations}. Clearly, our internal decision process goes beyond what we see- resolving ambiguity by assigning direction, intent and anthropomorphizing \cite{wan2021anthropomorphism}. One such so far neglected concept is the viewing direction of the animal, and not just where we look at as the supervisor. In this paper, we introduce three major contributions: First, we expose the existing methodologies of XAI by showing limitations: Highlighting pixels or areas alone, at least in the image domain, is not enough. Second, we introduce a new open-source optical illusion dataset featuring animal images labelled with bounding boxes, gaze and viewing direction annotations. And third, we demonstrate that integrating gaze direction and eye coordinates into the learning process improves model performance, even when all other aspects (architecture, epochs, learning rate, dataset structure, optimizer) remain the same. Our novel dataset is generated with sophisticated ChatGPT \cite{chatgpt} models and considerable computational power. It incorporates the understanding of a generative AI model in generating optical illusions while achieving images that are difficult for humans. However, we address how we mitigated potential biases in Section \ref{subsec:biasmit}. Furthermore, the difficulty of generating convincing optical illusions makes it hard to provide large datasets that are drawn by human artists: We argue that a machine generated set offers a valuable perspective into human vs AI perceptual learning.
\section{Related Work}
Explainable AI, often shortened to XAI, is broadly considered to be split into two categories: transparency design and post-hoc explanations \cite{xu2019explainable}. We criticize that the current approaches of visualization methods in the image domain highlight pixels or areas of images that show the importance of specific features. Saliency-based methods, such as LIME \cite{lime}, Grad-CAM \cite{selvaraju2017grad}, or Anchors \cite{ribeiro2018anchors} offer local explanations by attributing predictions to pixel importance or localized regions. The broader landscape of explainable AI encompasses countless methods \cite{lundberg2017unifiedapproachinterpretingmodel} from counterfactuals \cite{mothilal2020explaining, antoran2020getting}, prototypes \cite{nauta2023pip, DBLP:journals/corr/abs-1806-10574}, to global explainability methods \cite{setzu2021glocalx, morichetta2019explain, newen2022unsupervised}. We highlight the need for concept-based XAI in ambiguous settings. One type of concept-based XAI works is based on predefined concepts and relies on human supervision \cite{kim2018interpretability, bontempelli2022concept, yeh2020completeness, goyal2019explaining}. This is why this was extended to automatic concept-based extraction, which relies on segmentation strategies that then employ importance scores to dismiss outliers \cite{fel2023holistic, ghorbani2019towards, zhang2021invertible, fel2023craft}. However, this type of approach also comes with its limitations: We argue that all of these approaches still segment pixel-based logic. We apply ACE \cite{ghorbani2019towards} to our dataset as a prominent representative of the field and show with examples that it also had problems extracting useful concepts. We demonstrate in this work that general concepts for domains exist, such as gaze and the eye, that help the overall performance on a whole domain rather than just a specific type of animal. These concepts were not spotted when applying ACE. In our experimental section in Section \ref{sec:experiments}, as well as further experiments in the Appendix \ref{sec:Appendix}, we show the improvement when those concepts are considered versus learning without the concepts on our ambiguous dataset.

\begin{figure*}[h]
    \centering
    \includegraphics[width=11cm]{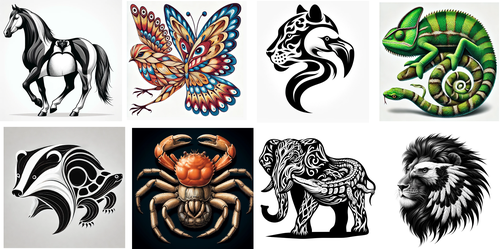}
    \caption{We feature here several examples of our dataset. For example, on the upper left side, a penguin can be seen hidden within a horse, depending on the direction we consider the animal to be looking. All of these examples have two animals distinguishable by the eye coordinate and the gaze vector, meaning they might be looking in the same direction, but their right eye (if more than one is visible) is positioned somewhere differently. That is why this is unique even in the case of the lion and eagle image, as the right eye of both is in a different position (image on the bottom right). For most of these images, however, the looking direction will also be distinguishable. The goal of this dataset was to test whether the gaze and eye coordinates prove to be useful general concepts in ambiguous settings but can, in general, be used for the evaluation of XAI as baseline for classification performance with optical illusions. More example pictures are included in the Appendix \ref{sec:Appendix} in Figure \ref{fig:overviewbirds}.}
    \label{fig:examples}
\end{figure*}
\subsection{Improving classifier performances using additional key features such as gaze annotations}
Apart from being beneficial in this specific example, gazes already prove useful to also enhance object detection model performances. It is important to note that we still provide a unique angle: For example, \cite{saab2019improving} take advantage of passively collected gaze information to decrease the number of training examples needed for the effective performance of a learner. In the literature, gaze annotations for humans often capture the direction of the person's gaze within the image. In contrast, for animals or inanimate objects, annotations tend to reflect points of interest identified by human observers, rather than attempting to label the gaze direction of the animal or object itself. We, however, focus on where the animal is looking and is being classified. There is various literature support that emphasizes that using additional features other than the image itself can create more effective learners with fewer training data: \cite{wang2017gaze} validate that gaze annotations can help improve the accuracies of classification approaches. However, both authors focused on annotating what humans look at and deem important. \cite{kellnhofer2019gaze360} use gaze annotations to increase the generalizability of their models compared to other benchmark datasets. 

Our work wants to incorporate the direction of the animal that this animate being looks in. \cite{wang2017gaze} evaluated their claims on two public datasets, and published their own dataset where food is annotated using important gaze points. The key difference between these ideas and our new contribution is that instead of focusing on saliency and repeating the concept of certain pixels or features that are most important in an image itself, we do not highlight where the person gazing at the image looks in but where the animate being contained in the image is looking. We give you an overview of related datasets and their contributions in Table \ref{tab:overview} in the Appendix for further research context clarification. 
\section{Limitations of current Explainable AI Algorithms}
\label{sec:limitations}
While current explainable AI algorithms do a phenomenal job at explaining attributions of models in general, our dataset is intentionally constructed to challenge local attribution methods by presenting overlapping or intermingled visual features from multiple animals. As we can see below, we provide some example areas where the highlighted features of the dataset mark features that do not belong to one of the animals specifically. In the following, we show instances where common XAI algorithms fail. We chose popular representatives from the saliency-based XAI area, such as Grad-CAM \cite{selvaraju2017grad} and integrated gradients. We then present an example using a prototypical XAI method, namely PipNet \cite{nauta2023pip}. We do the same for a representative of the automatic concept-based XAI area, ACE \cite{ghorbani2019towards}. We show that these methods fail to distinguish the two animals well due to the shared features of the animals: These limitations underscore a critical shortcoming of local attribution techniques and emphasize the need for concept-level reasoning.
\begin{figure}[h]
    \centering
    \includegraphics[width=\linewidth]{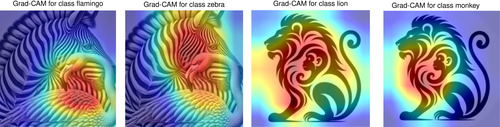}
    \caption{Pixel-based attribution explanations like Grad-CAM struggle to distinguish between the intermingling areas of the two animals. We later show that adding a single feature significantly enhances performance on ambiguous data.}
    \label{fig:gradcam-limitations}
\end{figure}

\begin{figure}[h]
    \centering
    \includegraphics[width=0.7\linewidth]{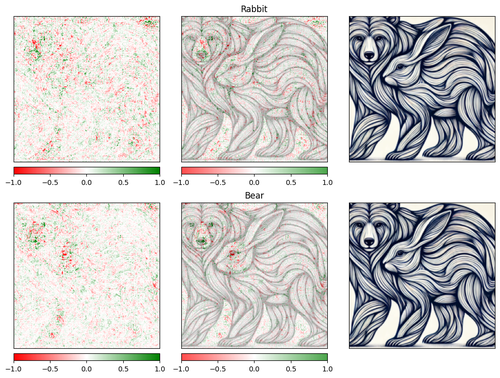}
    \caption{The same can be spotted for example using Integrated Gradients: The attributions for the classes are very similar, and often not very sensible. The explanation for bear clearly marks the rabbits' head in the picture. Both the rabbit and bear explanation include markings on the bear face area. Clearly, the model struggles to distinguish the two animals, and the explanations are limited in their meaningfulness and clarity.} %
    \label{fig:integrated_gradients_limitations}
\end{figure}
\begin{figure}[h]
    \centering
    \includegraphics[width=0.9\linewidth]{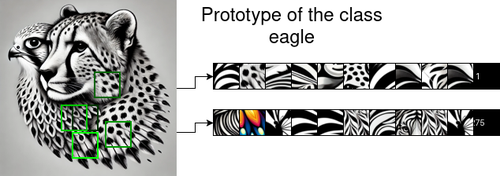}
    \caption{In this image, we see prototypes extracted via Pipnet \cite{nauta2023pip} for the eagle class. Pipnet also struggles to distinguish the cheetah fur and the eagle feathers. The darker the box, the more important it is for the classification. Again, we argue that this is due to the area-based and not concept-based explanations, a clear limitation for ambiguous data. Next to the original image, we see example concepts extracted that should show similar features.}
    \label{fig:Pipnetexample}
\end{figure}
\begin{figure}[h]
    \centering
    \includegraphics[width=0.9\linewidth]{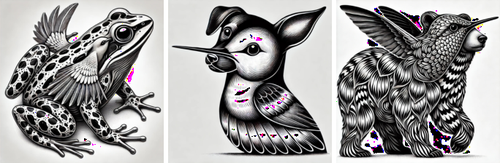}
    \caption{Example concepts extracted by ACE \cite{ghorbani2019towards} for the bird class. We argue that ACE cannot find abstract concepts, such as gaze direction, because it clusters segmentations on a pixel-based level. We argue that we are currently missing concept-based XAI that goes beyond the grouping of pixels.}
    \label{fig:birdACE}
\end{figure}

To go on, we show the same for Pipnet \cite{nauta2023pip}, a prototypical XAI algorithm. In Figure \ref{fig:Pipnetexample}, we can see that the prototypes marked by the algorithm contain all kinds of examples from animals, including eagle feathers (which were intentionally made similarly looking to confuse the learner). One of the prototypes even includes tiger eyes, if you look closely. The explanation of Pipnet marks both the cheetah fur and the eagle patterns because they admittedly look very similar. This proves our point, however: As a human, we draw in our head an invisible boundary between the cheetah and the eagle, because we think in concepts. The similarity of the fur does not matter to us. 
Furthermore, we then tried to evaluate how concept-based explainable AI methods, such as ACE \cite{ghorbani2019towards}, perform in the automatic detection of concepts in these ambiguous settings. While concept-based explanations in general require human-annotated concepts, ACE promises to automatically detect concepts using segmentation and clustering techniques using convolutional neural networks. When applying ACE to our dataset, it does not extract really meaningful concepts: For example, looking at the bird class in Figure \ref{fig:birdACE}, ACE does not really extract anything on the leftmost image; the middle one detected the dots on the bird wings, but this is not a meaningful feature in general. The rightmost image might be the wings with a lot of effort to see something, but it also highlights something meaningless at the bottom of the image. Furthermore, ACE fails to find the two general concepts that we suggest work well: gaze direction and the eyes.  We argue that this is because, again, ACE employs segmentation on a pixel level and derives concepts from there. This general setup is shared among automatic concept-detection methods \cite{fel2023holistic, zhang2021invertible, fel2023craft}. We argue beyond pixel-level segmentation. 

\section{Methodology}
One of the contributions of this paper is to show the usefulness of generalizable concepts such as eye coordinates and gaze direction. The direction of the gaze is a problem statement often considered in various different settings in literature \cite{mukherjee2015deep, liu2019differential}. Overall, it has already been recognised as an important factor in social interactions \cite{wang2021vision} or even in object detection itself \cite{bace2017facilitating}. The goal of this new dataset is to provide an ambiguous dataset baseline that might help evaluate future XAI algorithms on their ability to provide meaningful explanations on ambiguous data.

A key distinction for this proposition of including the gaze, in contrast to existing ones, is that we track where the animal or human is looking, not what the person looking at. We consider gaze following as: The gaze direction starting from the eye coordinate and then considering the head tilt, the direction of the gaze. The mathematical definition consists of the following components: 
\begin{itemize}
    \item \textbf{The Eye Position (e)}: A 2D point representing the eye's position in the plane, denoted as a vector $ [e_x, e_y] \in \mathbb{R}^2$.
    \item \textbf{Head Looking Direction(d):}: A unit vector representing the normalized direction in which the head is oriented, given as $[d_x,d_y] \in \mathbb{R}^2$. 
\end{itemize}
We can then define the gaze direction as $g=e+ \alpha \cdot d$, with $\alpha \in \mathbb{R}$ a scaling constant to ensure a set length. For explanation purposes, we aim to make it intuitively long enough to be well visible to the human eye. $\alpha$ is given a length based on the given image size. We normalize the gaze, ensuring a unified vector notation. For the sake of this evaluation methodology, we define that looking straight ahead is annotated as (0.0, 0.0). We always annotated the position of the right eye if two were present.
\section{Experiments: Benchmarking the concept of gaze direction and the eye on Ambivision- our dataset}
\label{sec:experiments}
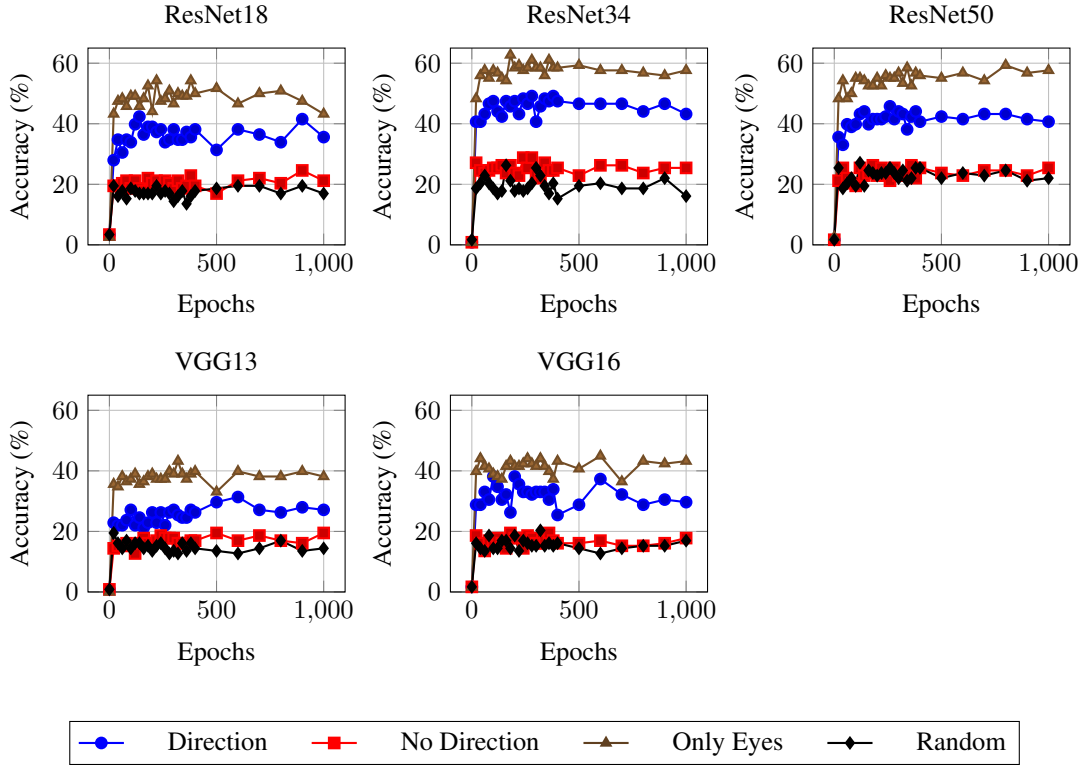
\begin{figure}[ht]
\centering
\begin{tikzpicture}
\begin{groupplot}[
    group style={
        group size=3 by 2,
        horizontal sep=1.4cm,
        vertical sep=2cm,
    },
    width=5cm,
    height=4.2cm,
    xlabel={Epochs},
    ylabel={Accuracy (\%)},
    grid=major,
    xtick={0, 500, 1000},
    ymin=0, ymax=65,
    legend style={font=\small},
    legend cell align={left},
]

\nextgroupplot[title={ResNet18}]
\addplot+[mark=*, thick] coordinates {(0,3.39)(20,27.97)(40,34.75)(60,30.51)(80,34.75)(100,33.90)(120,39.83)(140,42.37)(160,36.44)(180,38.98)(200,38.98)(220,37.29)(240,38.14)(260,33.90)(280,34.75)(300,38.14)(320,34.75)(340,34.75)(360,37.29)(380,35.59)(400,38.14)(500,31.36)(600,38.14)(700,36.44)(800,33.90)(900,41.53)(1000,35.59)};
\addplot+[mark=square*, thick] coordinates {(0,3.39)(20,19.49)(40,19.49)(60,20.34)(80,21.19)(100,18.64)(120,21.19)(140,18.64)(160,17.80)(180,22.03)(200,20.34)(220,21.19)(240,19.49)(260,21.19)(280,20.34)(300,18.64)(320,21.19)(340,18.64)(360,19.49)(380,22.88)(400,19.49)(500,16.95)(600,21.19)(700,22.03)(800,20.34)(900,24.58)(1000,21.19)};
\addplot+[mark=triangle*, thick] coordinates {(0,2.54)(20,43.22)(40,47.46)(60,48.31)(80,45.76)(100,49.15)(120,49.15)(140,45.76)(160,48.31)(180,52.54)(200,44.07)(220,54.24)(240,47.46)(260,48.31)(280,50.85)(300,46.61)(320,50.00)(340,49.15)(360,49.15)(380,54.24)(400,50.00)(500,51.69)(600,46.61)(700,50.00)(800,50.85)(900,47.46)(1000,43.22)};
\addplot+[mark=diamond*, thick] coordinates {(0,3.39)(20,19.49)(40,16.10)(60,17.80)(80,15.25)(100,18.64)(120,17.80)(140,16.95)(160,16.95)(180,16.95)(200,16.95)(220,19.49)(240,16.95)(260,17.80)(280,16.95)(300,14.41)(320,16.10)(340,17.80)(360,13.56)(380,16.10)(400,17.80)(500,18.64)(600,19.49)(700,19.49)(800,16.95)(900,19.49)(1000,16.95)};

\nextgroupplot[title={ResNet34}]
\addplot+[mark=*, thick] coordinates {(0,0.85)(20,40.68)(40,40.68)(60,43.22)(80,46.61)(100,47.46)(120,44.07)(140,42.37)(160,47.46)(180,45.76)(200,47.46)(220,43.22)(240,48.31)(260,46.61)(280,49.15)(300,40.68)(320,45.76)(340,48.31)(360,47.46)(380,49.15)(400,47.46)(500,46.61)(600,46.61)(700,46.61)(800,44.07)(900,46.61)(1000,43.22)};
\addplot+[mark=square*, thick] coordinates {(0,0.85)(20,27.12)(40,24.58)(60,23.73)(80,24.58)(100,25.42)(120,25.42)(140,26.27)(160,23.73)(180,26.27)(200,23.73)(220,22.88)(240,28.81)(260,25.42)(280,28.81)(300,22.88)(320,22.03)(340,27.12)(360,24.58)(380,24.58)(400,25.42)(500,22.88)(600,26.27)(700,26.27)(800,23.73)(900,25.42)(1000,25.42)};
\addplot+[mark=triangle*, thick] coordinates {(0,0.85)(20,48.31)(40,55.93)(60,57.63)(80,55.08)(100,57.63)(120,56.78)(140,55.08)(160,54.24)(180,62.71)(200,58.47)(220,59.32)(240,57.63)(260,58.47)(280,61.02)(300,58.47)(320,58.47)(340,55.93)(360,61.02)(380,58.47)(400,58.47)(500,59.32)(600,57.63)(700,57.63)(800,56.78)(900,55.93)(1000,57.63)};
\addplot+[mark=diamond*, thick] coordinates {(0,1.69)(20,18.64)(40,20.34)(60,22.88)(80,20.34)(100,18.64)(120,16.95)(140,17.80)(160,26.27)(180,21.19)(200,17.80)(220,18.64)(240,17.80)(260,18.64)(280,20.34)(300,25.42)(320,22.88)(340,19.49)(360,16.95)(380,20.34)(400,15.25)(500,19.49)(600,20.34)(700,18.64)(800,18.64)(900,22.03)(1000,16.10)};

\nextgroupplot[title={ResNet50}]
\addplot+[mark=*, thick] coordinates {(0,1.69)(20,35.59)(40,33.05)(60,39.83)(80,38.98)(100,39.83)(120,43.22)(140,44.07)(160,39.83)(180,41.53)(200,41.53)(220,41.53)(240,42.37)(260,45.76)(280,41.53)(300,44.07)(320,43.22)(340,38.14)(360,42.37)(380,44.07)(400,40.68)(500,42.37)(600,41.53)(700,43.22)(800,43.22)(900,41.53)(1000,40.68)};
\addplot+[mark=square*, thick] coordinates {(0,1.69)(20,21.19)(40,25.42)(60,22.03)(80,20.34)(100,19.49)(120,25.42)(140,22.88)(160,23.73)(180,26.27)(200,22.88)(220,25.42)(240,22.88)(260,21.19)(280,25.42)(300,22.88)(320,23.73)(340,22.88)(360,26.27)(380,22.03)(400,25.42)(500,23.73)(600,22.88)(700,24.58)(800,24.58)(900,22.88)(1000,25.42)};
\addplot+[mark=triangle*, thick] coordinates {(0,2.54)(20,48.31)(40,54.24)(60,48.31)(80,50.00)(100,55.08)(120,55.08)(140,54.24)(160,52.54)(180,52.54)(200,55.08)(220,52.54)(240,55.93)(260,55.08)(280,55.08)(300,56.78)(320,53.39)(340,58.47)(360,52.54)(380,56.78)(400,55.93)(500,55.08)(600,56.78)(700,54.24)(800,59.32)(900,56.78)(1000,57.63)};
\addplot+[mark=diamond*, thick] coordinates {(0,1.69)(20,25.42)(40,18.64)(60,20.34)(80,22.03)(100,19.49)(120,27.12)(140,19.49)(160,24.58)(180,23.73)(200,22.88)(220,23.73)(240,23.73)(260,25.42)(280,23.73)(300,22.03)(320,24.58)(340,21.19)(360,22.03)(380,25.42)(400,25.42)(500,22.03)(600,23.73)(700,22.88)(800,24.58)(900,21.19)(1000,22.03)};

\nextgroupplot[title={VGG13}]
\addplot+[mark=*, thick] coordinates {(0,0.85)(20,22.88)(40,22.03)(60,22.03)(80,23.73)(100,27.12)(120,22.03)(140,24.58)(160,21.19)(180,22.88)(200,26.27)(220,22.88)(240,26.27)(260,22.03)(280,26.27)(300,27.12)(320,25.42)(340,24.58)(360,24.58)(380,27.12)(400,26.27)(500,29.66)(600,31.36)(700,27.12)(800,26.27)(900,27.97)(1000,27.12)};
\addplot+[mark=square*, thick] coordinates {(0,0.85)(20,14.41)(40,15.25)(60,15.25)(80,16.10)(100,15.25)(120,12.71)(140,16.10)(160,17.80)(180,16.10)(200,16.10)(220,16.95)(240,18.64)(260,17.80)(280,17.80)(300,17.80)(320,16.10)(340,16.10)(360,16.10)(380,16.95)(400,16.95)(500,19.49)(600,16.95)(700,18.64)(800,16.95)(900,16.10)(1000,19.49)};
\addplot+[mark=triangle*, thick] coordinates {(0,0.00)(20,35.59)(40,34.75)(60,38.14)(80,36.44)(100,37.29)(120,38.98)(140,35.59)(160,36.44)(180,38.14)(200,38.98)(220,37.29)(240,37.29)(260,37.29)(280,39.83)(300,38.98)(320,43.22)(340,38.98)(360,37.29)(380,38.98)(400,39.83)(500,33.05)(600,39.83)(700,38.14)(800,38.14)(900,39.83)(1000,38.14)};
\addplot+[mark=diamond*, thick] coordinates {(0,0.85)(20,19.49)(40,16.10)(60,14.41)(80,16.95)(100,14.41)(120,16.10)(140,16.10)(160,14.41)(180,15.25)(200,13.56)(220,15.25)(240,16.10)(260,14.41)(280,12.71)(300,13.56)(320,12.71)(340,16.10)(360,13.56)(380,16.10)(400,14.41)(500,13.56)(600,12.71)(700,14.41)(800,16.95)(900,13.56)(1000,14.41)};

\nextgroupplot[title={VGG16}]
\addplot+[mark=*, thick] coordinates {(0,1.69)(20,28.81)(40,28.81)(60,33.05)(80,30.51)(100,38.14)(120,34.75)(140,30.51)(160,32.20)(180,26.27)(200,38.14)(220,35.59)(240,33.05)(260,33.05)(280,32.20)(300,33.05)(320,33.05)(340,33.05)(360,30.51)(380,33.90)(400,25.42)(500,28.81)(600,37.29)(700,32.20)(800,28.81)(900,30.51)(1000,29.66)};
\addplot+[mark=square*, thick] coordinates {(0,1.69)(20,18.64)(40,16.10)(60,13.56)(80,16.95)(100,15.25)(120,17.80)(140,16.10)(160,14.41)(180,19.49)(200,17.80)(220,17.80)(240,14.41)(260,18.64)(280,17.80)(300,16.10)(320,16.10)(340,16.95)(360,19.49)(380,16.95)(400,16.10)(500,16.10)(600,16.95)(700,15.25)(800,15.25)(900,16.10)(1000,17.80)};
\addplot+[mark=triangle*, thick] coordinates {(0,1.69)(20,39.83)(40,44.07)(60,41.53)(80,40.68)(100,38.98)(120,38.14)(140,37.29)(160,41.53)(180,43.22)(200,41.53)(220,41.53)(240,42.37)(260,44.07)(280,42.37)(300,41.53)(320,44.07)(340,41.53)(360,39.83)(380,37.29)(400,43.22)(500,40.68)(600,44.92)(700,36.44)(800,43.22)(900,42.37)(1000,43.22)};
\addplot+[mark=diamond*, thick] coordinates {(0,1.69)(20,16.10)(40,14.41)(60,13.56)(80,18.64)(100,14.41)(120,14.41)(140,16.95)(160,16.95)(180,14.41)(200,18.64)(220,13.56)(240,16.95)(260,16.10)(280,15.25)(300,15.25)(320,20.34)(340,15.25)(360,16.10)(380,15.25)(400,16.10)(500,14.41)(600,12.71)(700,14.41)(800,15.25)(900,15.25)(1000,16.95)};

\end{groupplot}
\path (current bounding box.south) ++(0,-0.5cm) node[anchor=north] {
    \begin{tikzpicture}
        \begin{axis}[
            hide axis,
            xmin=0, xmax=1,
            ymin=0, ymax=1,
            legend style={
                at={(0.5,1)},
                anchor=south,
                legend columns=4,
                column sep=0.5cm,
                /tikz/every even column/.append style={column sep=0.5cm}
            }
        ]
        \addlegendimage{color=blue, mark=*, thick}
        \addlegendentry{Direction}
        \addlegendimage{color=red, mark=square*, thick}
        \addlegendentry{No Direction}
        \addlegendimage{color=brown!50!black, mark=triangle*, thick}
        \addlegendentry{Only Eyes}
        \addlegendimage{color=black, mark=diamond*, thick}
        \addlegendentry{Random}

        \end{axis}
    \end{tikzpicture}
};
\end{tikzpicture}
\caption{Accuracy vs. Epochs at LR=0.0001 for ResNet and VGG models for direction, no direction, eye annotation and a baseline with random annotations. This clearly shows that the models, despite using the same architecture and hyperparameters, learned better with both the eye concept and viewing direction concept over all classes. A larger version of this plot is included in the Appendix \ref{sec:Appendix}.}
\label{fig:comparisonbaselines}
\end{figure}

Despite having literature supporting our claims that specific additional features help the learning process \cite{saab2019improving, wang2017gaze}, we wanted to provide an additional experimental evaluation on learning improvements when including the gaze vector. In this work, we focus not on the gaze of the observer (in contrast to eye-tracking literature), but on the depicted gaze direction of the object in the image—e.g., which way the animal is looking. This distinction is critical, as it provides a novel perspective. For evaluation purposes, we downloaded several state-of-the-art pre-trained Imagenet classifiers, namely Resnet18, Resnet34, Resnet52, VGG13 and VGG16 \cite{he2015deepresiduallearningimage}. We fine-tuned the networks on learning rates 0.0001, 0.00001, 0.000005 and over various amount of epochs. For the direction, we included arrows in the image that annotated the gaze direction in the training set, and validated it allowing either animal to be classified. We gathered the same results when only allowing one animal to be classified correctly. Figure \ref{fig:comparisonbaselines} shows the accuracies when allowing both classes. Despite allowing both classes as correct, in all cases, including the direction leads to significantly higher accuracies with otherwise the same hyperparameters. We include graphs with all tested learning rates in the Appendix \ref{sec:Appendix}. As baseline check, we tested the same by annotating random other areas in the image, to double check that the concept incorporated was meaningful. We also included one experiment where we tested how eye annotation alone performed in our dataset. Our experiments show that both gaze direction and eye annotation alone lead to meaningful improvements, raising accuracy rates by over 20\% in a setting allowing up to 1000 classes. Our results shown here were generated using the ADAM optimizer \cite{kingma2017adammethodstochasticoptimization}, because initial experiments revealed to us that this was the optimizer that produced the best accuracy results in practice. ADAM is known to be state-of-the-art in practice \cite{choi2019empirical}. All experiments were performed using a NVIDIA A100-SXM4-80GB GPU. Calculating lower epochs took seconds, the overall plotting of all accuracies up to 1.000 training epochs took several hours.

Marking the eye outperformed all other tests, however it should be noted that while the eye is a meaningful concept, this dataset did not necessarily insure uniqueness of looking direction, but always a unique combination of eye coordinate and gaze vector. The interesting part of this experiment was that this eye annotation did not have to be human-level accurate. A marking close to the eye sufficed to raise detection accuracies in ambiguous settings. An overview of those results can be seen in Figure \ref{fig:comparisonbaselines}. We evaluated two different approaches for the improvement of the optimization problem $\argmin_{g\in G}[L(f,g,\pi_{x'},e,d) + \Omega(g)]$: Our first approach was to optimize the task loss and then concatenate the gaze coordinates before evaluating the last softmax layer, then learn the data using a Multilayer Perceptron \cite{popescu2009multilayer}, essentially learning the following gaze information loss:
\begin{equation}
    L(f,g,\pi_{x'},e,d)= L\left( L_{task}(f,g, \pi_{x'}) \vert (e,d) \right)
\end{equation}

This particular setup, however, did not lead to a detectable increase in accuracy. We managed to increase the accuracy by including the gaze vector directly into the image information. %
However, we tested the accuracy improvement over several different types of architectures, using different amounts of epochs and learning rates, showing that the learning process can be significantly improved using the concepts eye and gaze direction. As can be seen in Figure \ref{fig:comparisonbaselines}, both eye coordinates and gaze direction led to significant improvements over the baseline of no annotation and random annotations. All other results for other learning rates can be found in the Appendix \ref{sec:Appendix}, but all show the same general results. 

\begin{figure}[h]
    \centering
    \includegraphics[width=8cm]{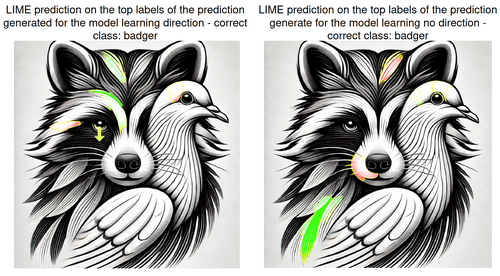}
    \caption{We see here that the LIME \cite{lime} explanation of the badger trained exactly the same as the other model with the exception of the direction vector is able to include fewer features shared by badger and pigeon, such as more area around the pigeon's eyes and where the fur of pigeon and badger overlap. The model also directly accounts for the eye gaze as well, which shows the usefulness of the feature in the learning process. We also made sure the gaze of the animal varied during the training. So, the animal cannot be classified simply because the arrow always points in the same direction.}
    \label{fig:badgerlime}
\end{figure}

 We preprocessed our dataset such that the shape of the images was adjusted to $3 \times 224 \times 224$ and normalized using the mean. Furthermore, we took a look at what exactly our network learns if we directly incorporate the direction arrow as well using LIME \cite{lime}: As you can see in Figure \ref{fig:badgerlime}, the network then highlights normal features but also the gaze vector as a feature, compared to worse features that overlap with the rivaling class and no highlighting of the eyes. We included another example of this as an interesting additional observation for Gradcam \cite{selvaraju2017grad} as well as for PipNet \cite{nauta2023pip} in the Appendix \ref{sec:Appendix}.
\section{Ambivision: Animal Optical Illusions- Our Dataset} In order to underline our problem statement, we include an entirely new self-generated dataset that was constructed using iterative prompt engineering with Chatgpt-4 and Chatgpt-4o \cite{chatgpt} of animal-based optical illusions. The dataset will be released as open-source with acceptance of the paper in combination with our evaluation scripts. Creating meaningful optical illusions is inherently difficult. First of all, note that we included psychological principles for good prompt engineering, such as Gestalt theory \cite{koffka2013principles}, which we elaborate together with an example prompt in our discussion of the bin s mitigation strategies in \ref{subsec:biasmit}. Each image in Ambivision depicts one animal hidden inside the body of another animal, creating an intentionally ambiguous perceptual boundary. The dataset consists of over 200 images annotated with the class label, eye coordinates, gaze vector and bounding boxes for both animals- providing rich, concept-level labels beyond standard object annotations. For every successful image generation, we had to enter around 150 prompts, resulting in roughly 30.000 ChatGPT prompts \cite{chatgpt}. Of the resulting dataset, 41 of those images are RGB, while the majority are black and white. This design choice reflects the increased difficulty in generating convincing color illusions. Ambivision is, to the best of our knowledge, the first dataset of its kind to systematically encode perceptual ambiguity with fine-grained concept annotations, making it a valuable baseline for evaluating both classification performance as well as explainability in ambiguous visual settings.
We provide four different versions of the dataset for the convenience of the user: We provide the dataset with the label of the animals, followed by the eye coordinates $(e_x,e_y)$, the normalized direction vector $(d_x,d_y)$ and the bounding boxes $(x_1,y_1)$, $(x_2,y_2)$. We provide the same dataset with direction arrows drawn directly into the image as a baseline for this work. An additional baseline is the dataset with random markings in the image and one baseline where just the eye is encircled.

\subsection{Bias mitigation strategies}
\label{subsec:biasmit}
Because this dataset was generated with ChatGPT \cite{chatgpt}, we had to make sure to ensure diverse results and mitigate biases. For each animal class, we ensured that the gaze and eye position vary. This applies equally to both animals in the image to avoid introducing unintentional class-specific cues. (In the Appendix \ref{sec:Appendix}, we include one exemplary overview of all eye positions and the spread of the eye positions for the bird class as plots). We varied artistic representation by prompting specifically for more or less realistic art styles. We specifically prompted the animals to be in alternating positions and poses, for example moving, sitting, flying, and eating. Another principle we applied for prompting is a principle borrowed from the psychological domain. The design rules for optical illusions go back to fundamental problems of psychology, such as Gestalt theory introduced as early as 1935 \cite{koffka2013principles}. These laws of conceptual organization give us an easy overview of what design concepts can trick our minds or make it difficult for our brains to correctly organize and interpret visual data. One such example is the concept of proximity: When something is in close proximity to something else, we are more likely to interpret it as belonging together than when they are further apart. The Gestalt principles provide us with useful guidelines for our prompts. Establishing this baseline dataset allows us to explore perception differences between AI learners and humans. One example of an initial bias that we mitigated was that we had immense trouble making sure illusions from the owl class did not always look straightforward. This might be due to the odd fact that owls can turn their neck 270 degrees in real life as well, which means that owls will, as a matter of fact, face you more often than not. By asking for a flying owl, for example, and asking specifically for different poses, it was, however, possible to get more diverse illusions. 

\paragraph{Prompt Example:} \textit{Generate an artistic black-and-white image featuring a fusion of a tiger and a falcon. The design blends the tiger's powerful stripes and muscular build with the falcon's sharp beak and impressive wingspan. This creates an optical illusion where, from one angle, it appears as a tiger crouching to pounce and, from another, as a falcon swooping down to capture its prey. The image emphasizes the shared features of predatory prowess and agility, highlighting the ferocity and grace of both animals.}

\section{Limitations and Future Work}
First of all, one major limitation is that the dataset is generated using Chatgpt, which is in itself vulnerable to internal biases. However, this is why we explicitly focused on addressing potential biases in our Section \ref{subsec:biasmit}. Furthermore, it is very hard to generate any kind of optical illusions, and our approach here allowed us to generate a dataset of over 200 images in a feasible manner. We also note that it is very interesting to have a dataset which is an optical illusion dataset for humans, but generated by a machine. Furthermore, in this dataset, we explicitly focused on two animals distinguishable by their gaze vector and eye coordinates.  Although this was outside the scope of evaluation, during our various and extensive prompt attempts, we generated other interesting concepts. Such as: examples of animals where more than one animal is hidden in the body of an animal, but also humans and animals in a mixed manner. We give access to the images that did not fulfil our criteria for evaluation purposes for exploratory and research purposes. Examples can be found in the Appendix \ref{sec:Appendix}. We were also able to generate interesting pictures where neither gaze nor eye coordinates was a distinguishing feature, but there were still two animals visible, again we include an example in the Appendix \ref{sec:Appendix}. Ultimately, this work invites a broader perspective: What if pixel-wise saliency is not the most effective approach to explainability in the image domain? What kinds of concepts should models truly be learning? Can we pursue more holistic learning strategies inspired by optical illusions and insights from cognitive psychology? These questions form a foundation for future exploration.

\section{Conclusion} In this paper, we challenge the prevailing paradigm in explainable AI (XAI) for visual data, which primarily revolves around pixel-based attributions and saliency maps. While such methods offer useful insights in many domains, they fall short when confronted with perceptual ambiguity—situations in which even human observers struggle to resolve competing interpretations. Inspired by classical optical illusions like the rabbit-duck example, we propose that meaningful explanations in these cases must go beyond pixels and capture abstract, semantic concepts such as gaze direction and eye position. While there have been some efforts in the concept-based domain, automatic generation of concepts again relies on pixel-based methodology and fails to capture concepts such as the viewing direction. To address this research gap, this paper introduces a novel dataset, Ambivision, presenting visually merged animal optical illusions. Each image is annotated with the animal classes, their right eye coordinate (if only one eye is visible, then that eye), the normalized viewing direction and bounding boxes for the animal. Through extensive experimentation across multiple state-of-the-art architectures and training regimes, we demonstrate that including such concept-level annotations—specifically gaze and eye location—leads to significant performance improvements on classification tasks in ambiguous settings. Additionally, we showed the limitations of popular existing XAI algorithms on this particular dataset due to its ambiguous nature. Ambivision represents a step toward rethinking how we evaluate and design explainability in AI. It opens up new directions for building more human-aligned explanations, ones that take into account not just what is seen but how it is perceived. We hope this work sparks new conversations about what it truly means to explain AI, as well as what concepts are best to incorporate into classification problems.

\begin{ack}
This work was supported by the Research Center Trustworthy Data Science and Security.
\end{ack}

\small
\bibliographystyle{plainnat}

\begin{thebibliography}{42}
\providecommand{\natexlab}[1]{#1}
\providecommand{\url}[1]{\texttt{#1}}
\expandafter\ifx\csname urlstyle\endcsname\relax
  \providecommand{\doi}[1]{doi: #1}\else
  \providecommand{\doi}{doi: \begingroup \urlstyle{rm}\Url}\fi

\bibitem[Antor{\'a}n et~al.(2020)Antor{\'a}n, Bhatt, Adel, Weller, and Hern{\'a}ndez-Lobato]{antoran2020getting}
Javier Antor{\'a}n, Umang Bhatt, Tameem Adel, Adrian Weller, and Jos{\'e}~Miguel Hern{\'a}ndez-Lobato.
\newblock Getting a clue: A method for explaining uncertainty estimates.
\newblock \emph{arXiv preprint arXiv:2006.06848}, 2020.

\bibitem[B{\^a}ce et~al.(2017)B{\^a}ce, Schlattner, Becker, and S{\"o}r{\"o}s]{bace2017facilitating}
Mihai B{\^a}ce, Philippe Schlattner, Vincent Becker, and G{\'a}bor S{\"o}r{\"o}s.
\newblock Facilitating object detection and recognition through eye gaze.
\newblock In \emph{19th International Conference on Human-Computer Interaction with Mobile Devices and Services (MobileHCI 2017)}. ETH Zurich, 2017.

\bibitem[Bontempelli et~al.(2022)Bontempelli, Teso, Tentori, Giunchiglia, and Passerini]{bontempelli2022concept}
Andrea Bontempelli, Stefano Teso, Katya Tentori, Fausto Giunchiglia, and Andrea Passerini.
\newblock Concept-level debugging of part-prototype networks.
\newblock \emph{arXiv preprint arXiv:2205.15769}, 2022.

\bibitem[Chen et~al.(2018)Chen, Li, Barnett, Su, and Rudin]{DBLP:journals/corr/abs-1806-10574}
Chaofan Chen, Oscar Li, Alina Barnett, Jonathan Su, and Cynthia Rudin.
\newblock This looks like that: deep learning for interpretable image recognition.
\newblock \emph{CoRR}, abs/1806.10574, 2018.
\newblock URL \url{http://arxiv.org/abs/1806.10574}.

\bibitem[Choi(2019)]{choi2019empirical}
D~Choi.
\newblock On empirical comparisons of optimizers for deep learning.
\newblock \emph{arXiv preprint arXiv:1910.05446}, 2019.

\bibitem[Fel et~al.(2023{\natexlab{a}})Fel, Boutin, B{\'e}thune, Cad{\`e}ne, Moayeri, And{\'e}ol, Chalvidal, and Serre]{fel2023holistic}
Thomas Fel, Victor Boutin, Louis B{\'e}thune, R{\'e}mi Cad{\`e}ne, Mazda Moayeri, L{\'e}o And{\'e}ol, Mathieu Chalvidal, and Thomas Serre.
\newblock A holistic approach to unifying automatic concept extraction and concept importance estimation.
\newblock \emph{Advances in Neural Information Processing Systems}, 36:\penalty0 54805--54818, 2023{\natexlab{a}}.

\bibitem[Fel et~al.(2023{\natexlab{b}})Fel, Picard, Bethune, Boissin, Vigouroux, Colin, Cad{\`e}ne, and Serre]{fel2023craft}
Thomas Fel, Agustin Picard, Louis Bethune, Thibaut Boissin, David Vigouroux, Julien Colin, R{\'e}mi Cad{\`e}ne, and Thomas Serre.
\newblock Craft: Concept recursive activation factorization for explainability.
\newblock In \emph{Proceedings of the IEEE/CVF Conference on Computer Vision and Pattern Recognition}, pages 2711--2721, 2023{\natexlab{b}}.

\bibitem[Funes~Mora et~al.(2014)Funes~Mora, Monay, and Odobez]{funes2014eyediap}
Kenneth~Alberto Funes~Mora, Florent Monay, and Jean-Marc Odobez.
\newblock Eyediap: A database for the development and evaluation of gaze estimation algorithms from rgb and rgb-d cameras.
\newblock In \emph{Proceedings of the symposium on eye tracking research and applications}, pages 255--258, 2014.

\bibitem[Ghorbani et~al.(2019)Ghorbani, Wexler, Zou, and Kim]{ghorbani2019towards}
Amirata Ghorbani, James Wexler, James~Y Zou, and Been Kim.
\newblock Towards automatic concept-based explanations.
\newblock \emph{Advances in neural information processing systems}, 32, 2019.

\bibitem[Gilani et~al.(2015)Gilani, Subramanian, Yan, Melcher, Sebe, and Winkler]{gilani2015pet}
Syed~Omer Gilani, Ramanathan Subramanian, Yan Yan, David Melcher, Nicu Sebe, and Stefan Winkler.
\newblock Pet: An eye-tracking dataset for animal-centric pascal object classes.
\newblock In \emph{2015 IEEE International Conference on Multimedia and Expo (ICME)}, pages 1--6. IEEE, 2015.

\bibitem[Goyal et~al.(2019)Goyal, Feder, Shalit, and Kim]{goyal2019explaining}
Yash Goyal, Amir Feder, Uri Shalit, and Been Kim.
\newblock Explaining classifiers with causal concept effect (cace).
\newblock \emph{arXiv preprint arXiv:1907.07165}, 2019.

\bibitem[He et~al.(2015)He, Zhang, Ren, and Sun]{he2015deepresiduallearningimage}
Kaiming He, Xiangyu Zhang, Shaoqing Ren, and Jian Sun.
\newblock Deep residual learning for image recognition, 2015.
\newblock URL \url{https://arxiv.org/abs/1512.03385}.

\bibitem[Kellnhofer et~al.(2019)Kellnhofer, Recasens, Stent, Matusik, and Torralba]{kellnhofer2019gaze360}
Petr Kellnhofer, Adria Recasens, Simon Stent, Wojciech Matusik, and Antonio Torralba.
\newblock Gaze360: Physically unconstrained gaze estimation in the wild.
\newblock In \emph{Proceedings of the IEEE/CVF international conference on computer vision}, pages 6912--6921, 2019.

\bibitem[Khan et~al.(2020)Khan, McDonagh, Khan, Shahabuddin, Arora, Khan, Shao, and Tzimiropoulos]{khan2020animalweb}
Muhammad~Haris Khan, John McDonagh, Salman Khan, Muhammad Shahabuddin, Aditya Arora, Fahad~Shahbaz Khan, Ling Shao, and Georgios Tzimiropoulos.
\newblock Animalweb: A large-scale hierarchical dataset of annotated animal faces.
\newblock In \emph{Proceedings of the IEEE/CVF conference on computer vision and pattern recognition}, pages 6939--6948, 2020.

\bibitem[Kim et~al.(2018)Kim, Wattenberg, Gilmer, Cai, Wexler, Viegas, et~al.]{kim2018interpretability}
Been Kim, Martin Wattenberg, Justin Gilmer, Carrie Cai, James Wexler, Fernanda Viegas, et~al.
\newblock Interpretability beyond feature attribution: Quantitative testing with concept activation vectors (tcav).
\newblock In \emph{International conference on machine learning}, pages 2668--2677. PMLR, 2018.

\bibitem[Kingma and Ba(2017)]{kingma2017adammethodstochasticoptimization}
Diederik~P. Kingma and Jimmy Ba.
\newblock Adam: A method for stochastic optimization, 2017.
\newblock URL \url{https://arxiv.org/abs/1412.6980}.

\bibitem[Koffka(2013)]{koffka2013principles}
Kurt Koffka.
\newblock \emph{Principles of Gestalt psychology}.
\newblock routledge, 2013.

\bibitem[Liu et~al.(2019)Liu, Yu, Mora, and Odobez]{liu2019differential}
Gang Liu, Yu~Yu, Kenneth A~Funes Mora, and Jean-Marc Odobez.
\newblock A differential approach for gaze estimation.
\newblock \emph{IEEE transactions on pattern analysis and machine intelligence}, 43\penalty0 (3):\penalty0 1092--1099, 2019.

\bibitem[Lundberg and Lee(2017{\natexlab{a}})]{lundberg2017unifiedapproachinterpretingmodel}
Scott Lundberg and Su-In Lee.
\newblock A unified approach to interpreting model predictions, 2017{\natexlab{a}}.
\newblock URL \url{https://arxiv.org/abs/1705.07874}.

\bibitem[Lundberg and Lee(2017{\natexlab{b}})]{lundberg2017unified}
Scott~M Lundberg and Su-In Lee.
\newblock A unified approach to interpreting model predictions.
\newblock \emph{Advances in neural information processing systems}, 30, 2017{\natexlab{b}}.

\bibitem[Morichetta et~al.(2019)Morichetta, Casas, and Mellia]{morichetta2019explain}
Andrea Morichetta, Pedro Casas, and Marco Mellia.
\newblock Explain-it: Towards explainable ai for unsupervised network traffic analysis.
\newblock In \emph{Proceedings of the 3rd ACM CoNEXT Workshop on Big DAta, Machine Learning and Artificial Intelligence for Data Communication Networks}, pages 22--28, 2019.

\bibitem[Mothilal et~al.(2020)Mothilal, Sharma, and Tan]{mothilal2020explaining}
Ramaravind~K Mothilal, Amit Sharma, and Chenhao Tan.
\newblock Explaining machine learning classifiers through diverse counterfactual explanations.
\newblock In \emph{Proceedings of the 2020 conference on fairness, accountability, and transparency}, pages 607--617, 2020.

\bibitem[Mukherjee and Robertson(2015)]{mukherjee2015deep}
Sankha~S Mukherjee and Neil~Martin Robertson.
\newblock Deep head pose: Gaze-direction estimation in multimodal video.
\newblock \emph{IEEE Transactions on Multimedia}, 17\penalty0 (11):\penalty0 2094--2107, 2015.

\bibitem[Nauta et~al.(2023)Nauta, Schl{\"o}tterer, Van~Keulen, and Seifert]{nauta2023pip}
Meike Nauta, J{\"o}rg Schl{\"o}tterer, Maurice Van~Keulen, and Christin Seifert.
\newblock Pip-net: Patch-based intuitive prototypes for interpretable image classification.
\newblock In \emph{Proceedings of the IEEE/CVF Conference on Computer Vision and Pattern Recognition}, pages 2744--2753, 2023.

\bibitem[Newen and M{\"u}ller(2022)]{newen2022unsupervised}
Carina Newen and Emmanuel M{\"u}ller.
\newblock Unsupervised deepview: Global explainability of uncertainties for high dimensional data.
\newblock In \emph{2022 IEEE International Conference on Knowledge Graph (ICKG)}, pages 196--202. IEEE, 2022.

\bibitem[Ng et~al.(2022)Ng, Ong, Zheng, Ni, Yeo, and Liu]{ng2022animal}
Xun~Long Ng, Kian~Eng Ong, Qichen Zheng, Yun Ni, Si~Yong Yeo, and Jun Liu.
\newblock Animal kingdom: A large and diverse dataset for animal behavior understanding.
\newblock In \emph{Proceedings of the IEEE/CVF conference on computer vision and pattern recognition}, pages 19023--19034, 2022.

\bibitem[OpenAI(2024)]{chatgpt}
OpenAI.
\newblock Chatgpt, 2024.
\newblock URL \url{https://www.openai.com/chatgpt}.
\newblock Retrieved from OpenAI.

\bibitem[Popescu et~al.(2009)Popescu, Balas, Perescu-Popescu, and Mastorakis]{popescu2009multilayer}
Marius-Constantin Popescu, Valentina~E Balas, Liliana Perescu-Popescu, and Nikos Mastorakis.
\newblock Multilayer perceptron and neural networks.
\newblock \emph{WSEAS Transactions on Circuits and Systems}, 8\penalty0 (7):\penalty0 579--588, 2009.

\bibitem[Ribeiro et~al.(2016)Ribeiro, Singh, and Guestrin]{lime}
Marco~Tulio Ribeiro, Sameer Singh, and Carlos Guestrin.
\newblock "why should i trust you?": Explaining the predictions of any classifier, 2016.

\bibitem[Ribeiro et~al.(2018)Ribeiro, Singh, and Guestrin]{ribeiro2018anchors}
Marco~Tulio Ribeiro, Sameer Singh, and Carlos Guestrin.
\newblock Anchors: High-precision model-agnostic explanations.
\newblock In \emph{Proceedings of the AAAI conference on artificial intelligence}, volume~32, 2018.

\bibitem[Saab et~al.(2019)Saab, Dunnmon, Ratner, Rubin, and R{\'e}]{saab2019improving}
Khaled Saab, Jared Dunnmon, Alexander Ratner, Daniel Rubin, and Christopher R{\'e}.
\newblock Improving sample complexity with observational supervision.
\newblock 2019.

\bibitem[Selvaraju et~al.(2017)Selvaraju, Cogswell, Das, Vedantam, Parikh, and Batra]{selvaraju2017grad}
Ramprasaath~R Selvaraju, Michael Cogswell, Abhishek Das, Ramakrishna Vedantam, Devi Parikh, and Dhruv Batra.
\newblock Grad-cam: Visual explanations from deep networks via gradient-based localization.
\newblock In \emph{Proceedings of the IEEE international conference on computer vision}, pages 618--626, 2017.

\bibitem[Setzu et~al.(2021)Setzu, Guidotti, Monreale, Turini, Pedreschi, and Giannotti]{setzu2021glocalx}
Mattia Setzu, Riccardo Guidotti, Anna Monreale, Franco Turini, Dino Pedreschi, and Fosca Giannotti.
\newblock Glocalx-from local to global explanations of black box ai models.
\newblock \emph{Artificial Intelligence}, 294:\penalty0 103457, 2021.

\bibitem[Sundararajan et~al.(2017)Sundararajan, Taly, and Yan]{sundararajan2017axiomatic}
Mukund Sundararajan, Ankur Taly, and Qiqi Yan.
\newblock Axiomatic attribution for deep networks.
\newblock In \emph{International conference on machine learning}, pages 3319--3328. PMLR, 2017.

\bibitem[Wan and Chen(2021)]{wan2021anthropomorphism}
Echo~Wen Wan and Rocky~Peng Chen.
\newblock Anthropomorphism and object attachment.
\newblock \emph{Current Opinion in Psychology}, 39:\penalty0 88--93, 2021.

\bibitem[Wang et~al.(2017)Wang, Thome, and Cord]{wang2017gaze}
Xin Wang, Nicolas Thome, and Matthieu Cord.
\newblock Gaze latent support vector machine for image classification improved by weakly supervised region selection.
\newblock \emph{Pattern Recognition}, 72:\penalty0 59--71, 2017.

\bibitem[Wang et~al.(2021)Wang, Zhang, Zhang, Zhao, and Liu]{wang2021vision}
Xinming Wang, Jianhua Zhang, Hanlin Zhang, Shuwen Zhao, and Honghai Liu.
\newblock Vision-based gaze estimation: A review.
\newblock \emph{IEEE Transactions on Cognitive and Developmental Systems}, 14\penalty0 (2):\penalty0 316--332, 2021.

\bibitem[Xu et~al.(2019)Xu, Uszkoreit, Du, Fan, Zhao, and Zhu]{xu2019explainable}
Feiyu Xu, Hans Uszkoreit, Yangzhou Du, Wei Fan, Dongyan Zhao, and Jun Zhu.
\newblock Explainable ai: A brief survey on history, research areas, approaches and challenges.
\newblock In \emph{Natural Language Processing and Chinese Computing: 8th CCF International Conference, NLPCC 2019, Dunhuang, China, October 9--14, 2019, Proceedings, Part II 8}, pages 563--574. Springer, 2019.

\bibitem[Yeh et~al.(2020)Yeh, Kim, Arik, Li, Pfister, and Ravikumar]{yeh2020completeness}
Chih-Kuan Yeh, Been Kim, Sercan Arik, Chun-Liang Li, Tomas Pfister, and Pradeep Ravikumar.
\newblock On completeness-aware concept-based explanations in deep neural networks.
\newblock \emph{Advances in neural information processing systems}, 33:\penalty0 20554--20565, 2020.

\bibitem[Zhang et~al.(2021)Zhang, Madumal, Miller, Ehinger, and Rubinstein]{zhang2021invertible}
Ruihan Zhang, Prashan Madumal, Tim Miller, Krista~A Ehinger, and Benjamin~IP Rubinstein.
\newblock Invertible concept-based explanations for cnn models with non-negative concept activation vectors.
\newblock In \emph{Proceedings of the AAAI Conference on Artificial Intelligence}, volume~35, pages 11682--11690, 2021.

\bibitem[Zhang et~al.(2015)Zhang, Sugano, Fritz, and Bulling]{zhang2015appearance}
Xucong Zhang, Yusuke Sugano, Mario Fritz, and Andreas Bulling.
\newblock Appearance-based gaze estimation in the wild.
\newblock In \emph{Proceedings of the IEEE conference on computer vision and pattern recognition}, pages 4511--4520, 2015.

\bibitem[Zhang et~al.(2020)Zhang, Park, Beeler, Bradley, Tang, and Hilliges]{zhang2020eth}
Xucong Zhang, Seonwook Park, Thabo Beeler, Derek Bradley, Siyu Tang, and Otmar Hilliges.
\newblock Eth-xgaze: A large scale dataset for gaze estimation under extreme head pose and gaze variation.
\newblock In \emph{Computer Vision--ECCV 2020: 16th European Conference, Glasgow, UK, August 23--28, 2020, Proceedings, Part V 16}, pages 365--381. Springer, 2020.

\end{thebibliography}

\newpage
\appendix

\section{Technical Appendices and Supplementary Material}
\label{sec:Appendix}
We promised to deliver additional material for various points in the paper for whomever it may interest. First, we show an exemplary overview of existing types of gaze annotation datasets and their purposes for a literature overview in Table \ref{tab:overview}. In the following, we show the spread of the eye coordinates of the class \enquote{bird} and then a scatter plot of all eye coordinates in total.
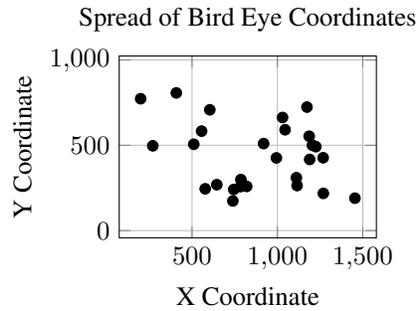
\begin{figure}[h]
\centering
\begin{tikzpicture}
  \begin{axis}[
    title={Spread of Bird Eye Coordinates},
    xlabel={X Coordinate},
    ylabel={Y Coordinate},
    width=5cm,
    height=4cm,
    grid=both,
    axis equal,
    enlargelimits=true,
    scatter/classes={a={mark=*,blue}},
    ]
    \addplot[only marks, scatter, mark size=2pt]
    coordinates {
      (784,258) (739,174) (199,773) (1173,724)
      (645,269) (1225,492) (1205,500) (994,426)
      (1189,417) (1454,190) (270,497) (510,506)
      (1268,427) (744,241) (577,245) (919,510)
      (1186,553) (1045,591) (820,259) (1115,264)
      (408,807) (1269,218) (604,708) (556,583)
      (1030,663) (786,299) (1111,310)
    };
  \end{axis}
\end{tikzpicture}
\caption{Scatter plot showing the distribution of bird eye coordinates.}
\label{fig:eye-coordinates}
\end{figure}

\begin{figure}[h]
    \centering
    \includegraphics[width=0.6\linewidth]{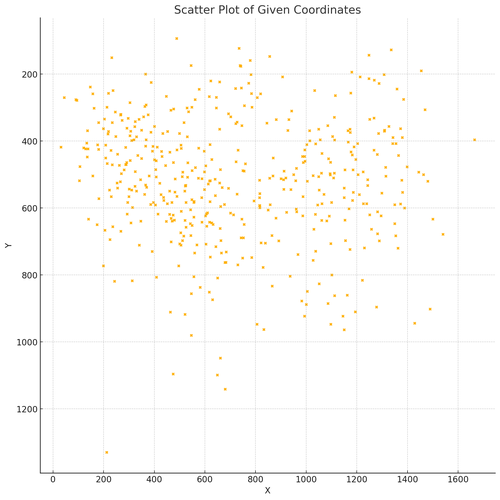}
    \caption{This scatter plot shows the spread of all eye coordinates from all classes. It is pretty diverse, considering the eyes are placed within the body of the animal and can therefore not occur on the immediate sides.}
    \label{fig:collected_eyes}
\end{figure}

\begin{figure}[H]
    \centering
    \includegraphics[width=10cm]{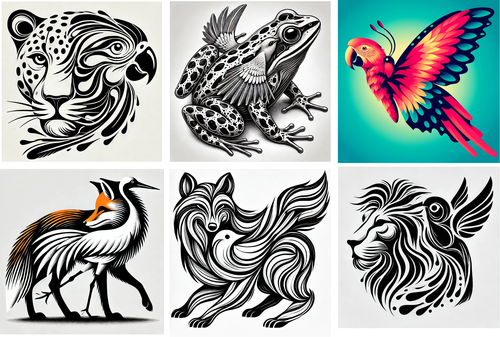}
    \caption{An overview of example instances from the \enquote{bird} class to demonstrate the diversity of the illusions}
    \label{fig:overviewbirds}
\end{figure}

To show that the eye coordinates in general are not placed in one area, we show here the distribution of all eye coordinates in one collected plot in Figure \ref{fig:collected_eyes}. The only tendency visible is that more of them are in the center areas than the edges, which makes sense because the body of an animal is always around the eye. 
\begin{table*}[tbp]

  \caption{ An exemplary Overview of existing Types of Gaze Annotation Datasets and their purposes.}
  \scalebox{0.52}{%

  \label{tab:overview}
  \centering
  \begin{tblr}{
      colspec={l|l|l},
      row{1}={font=\bfseries},
      column{1}={font=\itshape},
      row{1,3,4,7,9,10,13,15,16,17,18}={bg=gray!10},
    }   
    \hline
    \hline
    \textit{\textbf{Dataset Name}} & \textit{Type of Data} & Type of Annotation  \\
    \hline
     \textbf{Cathegory: Human Gaze Annotations } & &  \\
      \hline
    1. MPIIGaze \cite{zhang2015appearance} & large-scale dataset of Human Images of 15 subjects & Focus on different lightings, variable places and  \\
    &  & day times, as natural settings as possible\\
    \hline
      2. EyeDiap \cite{funes2014eyediap} & human gaze estimation dataset of 16 test subjects & head pose variations, gaze poses ground truth given   \\
     & changes in ambient and sensing conditions & by the 3d poses of the visual target  \\
     \hline
     3. ETH-XGaze \cite{zhang2020eth} & human gaze estimation dataset of 110 participants & head pose variations and different lighting \\
     \hline
      4. Gaze360  \cite{kellnhofer2019gaze360} & human gaze estimation dataset of 238 subjects & wide range of lighting conditions \\

    \hline
    \textbf{Cathegory: Animal Gaze Annotations } & &  \\
      \hline
      5. Animal Kingdom Dataset \cite{ng2022animal} & Video Dataset of Animals  & annotated for relevant animal behavior, \\
    && no eye gaze annotation but pose estimation \\
    \hline
    6. PET \cite{gilani2015pet} & classes bird, cat, cow, dog, horse, and sheep  & eye movements recorded of human points of interest \\
    & tracking eye position for visual points of importance from 40 users &  for free vision task and visual search\\
    \hline
     7. AnimalWeb \cite{khan2020animalweb} & animal faces collected from 350 species & annotated with 9 land-marks on key facial features \\
    \hline
    \textbf{Our Dataset: Ambivision} & Drawn Animal Images generated by ChatGPT4 and ChatGPT4o & Gaze vector in 2D format, animal labels for both animals\\
    \textbf{Optical Illusions)}  & consisting of always two animals, one merged/hidden in the other animal & animals were distinguishable by their right eye coordinate and \\
    && normalized direction vector. If it stared straight ahead, gaze \\
    &&vector was assigned as 0.0, 0.0. annotated bounding boxes \\
    \hline
    \hline
  \end{tblr}}
\end{table*} 
\newpage
 In the following, we also include more example images of the dataset to demonstrate the quality and diversity of the illusions in Figure \ref{fig:overviewbirds}.
\newpage
To go on, we continue by including all experimental results for different learning rates and architectures in larger format in this Appendix to demonstrate the validity of our results. 
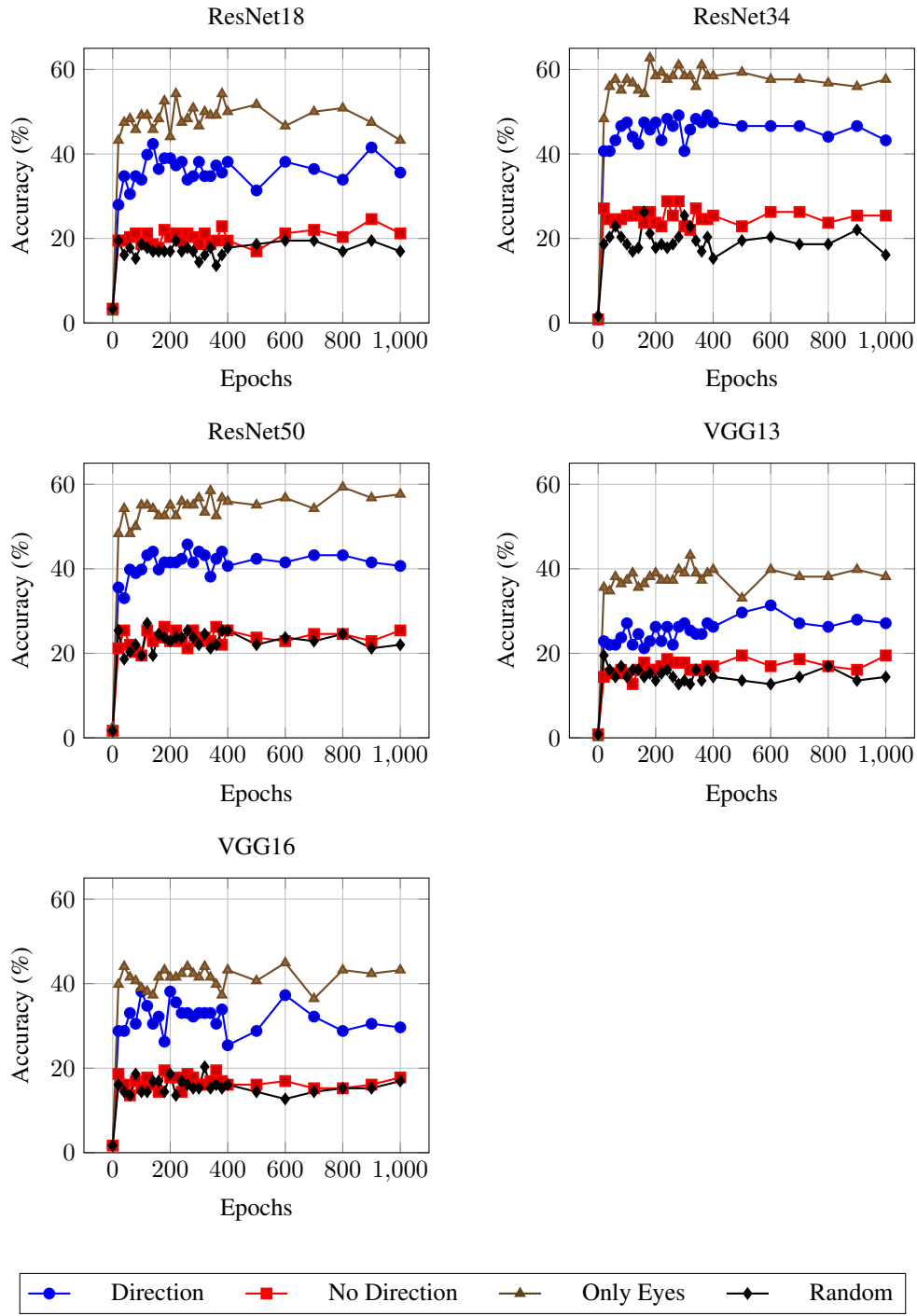
\begin{figure}[ht]
\centering
\begin{tikzpicture}
\begin{groupplot}[
    group style={
        group size=2 by 3,
        horizontal sep=2cm,
        vertical sep=2cm,
    },
    width=6.5cm,
    height=5.5cm,
    xlabel={Epochs},
    ylabel={Accuracy (\%)},
    grid=major,
    xtick={0, 200, 400, 600, 800, 1000},
    ymin=0, ymax=65,
    legend style={font=\small},
    legend cell align={left},
]

\nextgroupplot[title={ResNet18}]
\addplot+[mark=*, thick] coordinates {(0,3.39)(20,27.97)(40,34.75)(60,30.51)(80,34.75)(100,33.90)(120,39.83)(140,42.37)(160,36.44)(180,38.98)(200,38.98)(220,37.29)(240,38.14)(260,33.90)(280,34.75)(300,38.14)(320,34.75)(340,34.75)(360,37.29)(380,35.59)(400,38.14)(500,31.36)(600,38.14)(700,36.44)(800,33.90)(900,41.53)(1000,35.59)};
\addplot+[mark=square*, thick] coordinates {(0,3.39)(20,19.49)(40,19.49)(60,20.34)(80,21.19)(100,18.64)(120,21.19)(140,18.64)(160,17.80)(180,22.03)(200,20.34)(220,21.19)(240,19.49)(260,21.19)(280,20.34)(300,18.64)(320,21.19)(340,18.64)(360,19.49)(380,22.88)(400,19.49)(500,16.95)(600,21.19)(700,22.03)(800,20.34)(900,24.58)(1000,21.19)};
\addplot+[mark=triangle*, thick] coordinates {(0,2.54)(20,43.22)(40,47.46)(60,48.31)(80,45.76)(100,49.15)(120,49.15)(140,45.76)(160,48.31)(180,52.54)(200,44.07)(220,54.24)(240,47.46)(260,48.31)(280,50.85)(300,46.61)(320,50.00)(340,49.15)(360,49.15)(380,54.24)(400,50.00)(500,51.69)(600,46.61)(700,50.00)(800,50.85)(900,47.46)(1000,43.22)};
\addplot+[mark=diamond*, thick] coordinates {(0,3.39)(20,19.49)(40,16.10)(60,17.80)(80,15.25)(100,18.64)(120,17.80)(140,16.95)(160,16.95)(180,16.95)(200,16.95)(220,19.49)(240,16.95)(260,17.80)(280,16.95)(300,14.41)(320,16.10)(340,17.80)(360,13.56)(380,16.10)(400,17.80)(500,18.64)(600,19.49)(700,19.49)(800,16.95)(900,19.49)(1000,16.95)};

\nextgroupplot[title={ResNet34}]
\addplot+[mark=*, thick] coordinates {(0,0.85)(20,40.68)(40,40.68)(60,43.22)(80,46.61)(100,47.46)(120,44.07)(140,42.37)(160,47.46)(180,45.76)(200,47.46)(220,43.22)(240,48.31)(260,46.61)(280,49.15)(300,40.68)(320,45.76)(340,48.31)(360,47.46)(380,49.15)(400,47.46)(500,46.61)(600,46.61)(700,46.61)(800,44.07)(900,46.61)(1000,43.22)};
\addplot+[mark=square*, thick] coordinates {(0,0.85)(20,27.12)(40,24.58)(60,23.73)(80,24.58)(100,25.42)(120,25.42)(140,26.27)(160,23.73)(180,26.27)(200,23.73)(220,22.88)(240,28.81)(260,25.42)(280,28.81)(300,22.88)(320,22.03)(340,27.12)(360,24.58)(380,24.58)(400,25.42)(500,22.88)(600,26.27)(700,26.27)(800,23.73)(900,25.42)(1000,25.42)};
\addplot+[mark=triangle*, thick] coordinates {(0,0.85)(20,48.31)(40,55.93)(60,57.63)(80,55.08)(100,57.63)(120,56.78)(140,55.08)(160,54.24)(180,62.71)(200,58.47)(220,59.32)(240,57.63)(260,58.47)(280,61.02)(300,58.47)(320,58.47)(340,55.93)(360,61.02)(380,58.47)(400,58.47)(500,59.32)(600,57.63)(700,57.63)(800,56.78)(900,55.93)(1000,57.63)};
\addplot+[mark=diamond*, thick] coordinates {(0,1.69)(20,18.64)(40,20.34)(60,22.88)(80,20.34)(100,18.64)(120,16.95)(140,17.80)(160,26.27)(180,21.19)(200,17.80)(220,18.64)(240,17.80)(260,18.64)(280,20.34)(300,25.42)(320,22.88)(340,19.49)(360,16.95)(380,20.34)(400,15.25)(500,19.49)(600,20.34)(700,18.64)(800,18.64)(900,22.03)(1000,16.10)};

\nextgroupplot[title={ResNet50}]
\addplot+[mark=*, thick] coordinates {(0,1.69)(20,35.59)(40,33.05)(60,39.83)(80,38.98)(100,39.83)(120,43.22)(140,44.07)(160,39.83)(180,41.53)(200,41.53)(220,41.53)(240,42.37)(260,45.76)(280,41.53)(300,44.07)(320,43.22)(340,38.14)(360,42.37)(380,44.07)(400,40.68)(500,42.37)(600,41.53)(700,43.22)(800,43.22)(900,41.53)(1000,40.68)};
\addplot+[mark=square*, thick] coordinates {(0,1.69)(20,21.19)(40,25.42)(60,22.03)(80,20.34)(100,19.49)(120,25.42)(140,22.88)(160,23.73)(180,26.27)(200,22.88)(220,25.42)(240,22.88)(260,21.19)(280,25.42)(300,22.88)(320,23.73)(340,22.88)(360,26.27)(380,22.03)(400,25.42)(500,23.73)(600,22.88)(700,24.58)(800,24.58)(900,22.88)(1000,25.42)};
\addplot+[mark=triangle*, thick] coordinates {(0,2.54)(20,48.31)(40,54.24)(60,48.31)(80,50.00)(100,55.08)(120,55.08)(140,54.24)(160,52.54)(180,52.54)(200,55.08)(220,52.54)(240,55.93)(260,55.08)(280,55.08)(300,56.78)(320,53.39)(340,58.47)(360,52.54)(380,56.78)(400,55.93)(500,55.08)(600,56.78)(700,54.24)(800,59.32)(900,56.78)(1000,57.63)};
\addplot+[mark=diamond*, thick] coordinates {(0,1.69)(20,25.42)(40,18.64)(60,20.34)(80,22.03)(100,19.49)(120,27.12)(140,19.49)(160,24.58)(180,23.73)(200,22.88)(220,23.73)(240,23.73)(260,25.42)(280,23.73)(300,22.03)(320,24.58)(340,21.19)(360,22.03)(380,25.42)(400,25.42)(500,22.03)(600,23.73)(700,22.88)(800,24.58)(900,21.19)(1000,22.03)};

\nextgroupplot[title={VGG13}]
\addplot+[mark=*, thick] coordinates {(0,0.85)(20,22.88)(40,22.03)(60,22.03)(80,23.73)(100,27.12)(120,22.03)(140,24.58)(160,21.19)(180,22.88)(200,26.27)(220,22.88)(240,26.27)(260,22.03)(280,26.27)(300,27.12)(320,25.42)(340,24.58)(360,24.58)(380,27.12)(400,26.27)(500,29.66)(600,31.36)(700,27.12)(800,26.27)(900,27.97)(1000,27.12)};
\addplot+[mark=square*, thick] coordinates {(0,0.85)(20,14.41)(40,15.25)(60,15.25)(80,16.10)(100,15.25)(120,12.71)(140,16.10)(160,17.80)(180,16.10)(200,16.10)(220,16.95)(240,18.64)(260,17.80)(280,17.80)(300,17.80)(320,16.10)(340,16.10)(360,16.10)(380,16.95)(400,16.95)(500,19.49)(600,16.95)(700,18.64)(800,16.95)(900,16.10)(1000,19.49)};
\addplot+[mark=triangle*, thick] coordinates {(0,0.00)(20,35.59)(40,34.75)(60,38.14)(80,36.44)(100,37.29)(120,38.98)(140,35.59)(160,36.44)(180,38.14)(200,38.98)(220,37.29)(240,37.29)(260,37.29)(280,39.83)(300,38.98)(320,43.22)(340,38.98)(360,37.29)(380,38.98)(400,39.83)(500,33.05)(600,39.83)(700,38.14)(800,38.14)(900,39.83)(1000,38.14)};
\addplot+[mark=diamond*, thick] coordinates {(0,0.85)(20,19.49)(40,16.10)(60,14.41)(80,16.95)(100,14.41)(120,16.10)(140,16.10)(160,14.41)(180,15.25)(200,13.56)(220,15.25)(240,16.10)(260,14.41)(280,12.71)(300,13.56)(320,12.71)(340,16.10)(360,13.56)(380,16.10)(400,14.41)(500,13.56)(600,12.71)(700,14.41)(800,16.95)(900,13.56)(1000,14.41)};

\nextgroupplot[title={VGG16}]
\addplot+[mark=*, thick] coordinates {(0,1.69)(20,28.81)(40,28.81)(60,33.05)(80,30.51)(100,38.14)(120,34.75)(140,30.51)(160,32.20)(180,26.27)(200,38.14)(220,35.59)(240,33.05)(260,33.05)(280,32.20)(300,33.05)(320,33.05)(340,33.05)(360,30.51)(380,33.90)(400,25.42)(500,28.81)(600,37.29)(700,32.20)(800,28.81)(900,30.51)(1000,29.66)};
\addplot+[mark=square*, thick] coordinates {(0,1.69)(20,18.64)(40,16.10)(60,13.56)(80,16.95)(100,15.25)(120,17.80)(140,16.10)(160,14.41)(180,19.49)(200,17.80)(220,17.80)(240,14.41)(260,18.64)(280,17.80)(300,16.10)(320,16.10)(340,16.95)(360,19.49)(380,16.95)(400,16.10)(500,16.10)(600,16.95)(700,15.25)(800,15.25)(900,16.10)(1000,17.80)};
\addplot+[mark=triangle*, thick] coordinates {(0,1.69)(20,39.83)(40,44.07)(60,41.53)(80,40.68)(100,38.98)(120,38.14)(140,37.29)(160,41.53)(180,43.22)(200,41.53)(220,41.53)(240,42.37)(260,44.07)(280,42.37)(300,41.53)(320,44.07)(340,41.53)(360,39.83)(380,37.29)(400,43.22)(500,40.68)(600,44.92)(700,36.44)(800,43.22)(900,42.37)(1000,43.22)};
\addplot+[mark=diamond*, thick] coordinates {(0,1.69)(20,16.10)(40,14.41)(60,13.56)(80,18.64)(100,14.41)(120,14.41)(140,16.95)(160,16.95)(180,14.41)(200,18.64)(220,13.56)(240,16.95)(260,16.10)(280,15.25)(300,15.25)(320,20.34)(340,15.25)(360,16.10)(380,15.25)(400,16.10)(500,14.41)(600,12.71)(700,14.41)(800,15.25)(900,15.25)(1000,16.95)};

\end{groupplot}
\path (current bounding box.south) ++(0,-0.5cm) node[anchor=north] {
    \begin{tikzpicture}
        \begin{axis}[
            hide axis,
            xmin=0, xmax=1,
            ymin=0, ymax=1,
            legend style={
                at={(0.5,1)},
                anchor=south,
                legend columns=4,
                column sep=0.5cm,
                /tikz/every even column/.append style={column sep=0.5cm}
            }
        ]
        \addlegendimage{color=blue, mark=*, thick}
        \addlegendentry{Direction}
        \addlegendimage{color=red, mark=square*, thick}
        \addlegendentry{No Direction}
        \addlegendimage{color=brown!50!black, mark=triangle*, thick}
        \addlegendentry{Only Eyes}
        \addlegendimage{color=black, mark=diamond*, thick}
        \addlegendentry{Random}

        \end{axis}
    \end{tikzpicture}
};
\end{tikzpicture}
\caption{Accuracy vs. Epochs at LR=0.0001 for ResNet and VGG models across different annotation conditions.}
\end{figure}

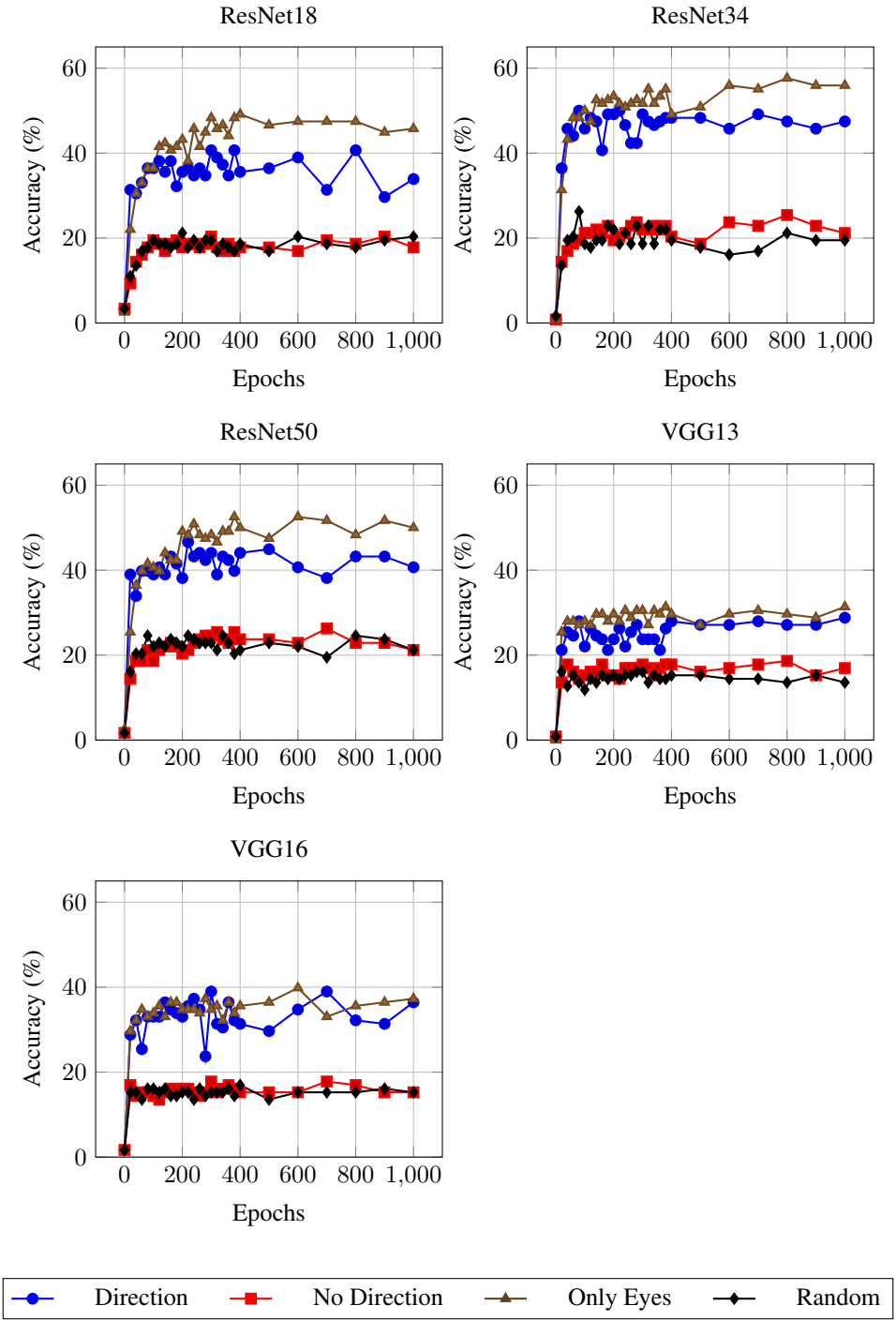
\begin{figure}[H]
\centering
\begin{tikzpicture}
\begin{groupplot}[
    group style={
        group size=2 by 3,
        horizontal sep=1.2cm,
        vertical sep=2cm,
    },
    width=6.5cm,
    height=5.5cm,
    xlabel={Epochs},
    ylabel={Accuracy (\%)},
    grid=major,
    xtick={0, 200, 400, 600, 800, 1000},
    ymin=0, ymax=65,
    legend style={font=\small},
    legend cell align={left},
]

\nextgroupplot[title={ResNet18}]
\addplot+[mark=*, thick] coordinates {(0,3.39)(20,31.36)(40,30.51)(60,33.05)(80,36.44)(100,36.44)(120,38.14)(140,35.59)(160,38.14)(180,32.20)(200,35.59)(220,36.44)(240,34.75)(260,36.44)(280,34.75)(300,40.68)(320,38.98)(340,37.29)(360,34.75)(380,40.68)(400,35.59)(500,36.44)(600,38.98)(700,31.36)(800,40.68)(900,29.66)(1000,33.90)};
\addplot+[mark=square*, thick] coordinates {(0,3.39)(20,9.32)(40,14.41)(60,16.10)(80,17.80)(100,19.49)(120,18.64)(140,16.95)(160,18.64)(180,19.49)(200,17.80)(220,18.64)(240,18.64)(260,17.80)(280,18.64)(300,20.34)(320,17.80)(340,16.95)(360,18.64)(380,16.95)(400,17.80)(500,17.80)(600,16.95)(700,19.49)(800,18.64)(900,20.34)(1000,17.80)};
\addplot+[mark=triangle*, thick] coordinates {(0,2.54)(20,22.03)(40,30.51)(60,33.05)(80,36.44)(100,36.44)(120,41.53)(140,42.37)(160,40.68)(180,41.53)(200,43.22)(220,38.14)(240,45.76)(260,41.53)(280,44.92)(300,48.31)(320,45.76)(340,46.61)(360,44.07)(380,48.31)(400,49.15)(500,46.61)(600,47.46)(700,47.46)(800,47.46)(900,44.92)(1000,45.76)};
\addplot+[mark=diamond*, thick] coordinates {(0,3.39)(20,11.02)(40,13.56)(60,16.95)(80,17.80)(100,19.49)(120,18.64)(140,18.64)(160,17.80)(180,18.64)(200,21.19)(220,17.80)(240,19.49)(260,17.80)(280,19.49)(300,19.49)(320,16.95)(340,18.64)(360,17.80)(380,16.95)(400,18.64)(500,16.95)(600,20.34)(700,18.64)(800,17.80)(900,19.49)(1000,20.34)};

\nextgroupplot[title={ResNet34}]
\addplot+[mark=*, thick] coordinates {(0,0.85)(20,36.44)(40,45.76)(60,44.07)(80,50.00)(100,45.76)(120,48.31)(140,47.46)(160,40.68)(180,49.15)(200,49.15)(220,50.00)(240,46.61)(260,42.37)(280,42.37)(300,49.15)(320,47.46)(340,46.61)(360,47.46)(380,48.31)(400,48.31)(500,48.31)(600,45.76)(700,49.15)(800,47.46)(900,45.76)(1000,47.46)};
\addplot+[mark=square*, thick] coordinates {(0,0.85)(20,14.41)(40,16.95)(60,18.64)(80,19.49)(100,21.19)(120,21.19)(140,22.03)(160,21.19)(180,22.88)(200,19.49)(220,20.34)(240,21.19)(260,22.88)(280,23.73)(300,22.03)(320,22.03)(340,22.88)(360,22.03)(380,22.88)(400,20.34)(500,18.64)(600,23.73)(700,22.88)(800,25.42)(900,22.88)(1000,21.19)};
\addplot+[mark=triangle*, thick] coordinates {(0,0.85)(20,31.36)(40,43.22)(60,48.31)(80,48.31)(100,50.00)(120,47.46)(140,52.54)(160,51.69)(180,52.54)(200,53.39)(220,51.69)(240,50.85)(260,51.69)(280,52.54)(300,51.69)(320,55.08)(340,51.69)(360,53.39)(380,55.08)(400,49.15)(500,50.85)(600,55.93)(700,55.08)(800,57.63)(900,55.93)(1000,55.93)};
\addplot+[mark=diamond*, thick] coordinates {(0,1.69)(20,13.56)(40,19.49)(60,20.34)(80,26.27)(100,18.64)(120,17.80)(140,19.49)(160,19.49)(180,22.88)(200,22.03)(220,18.64)(240,21.19)(260,18.64)(280,22.88)(300,18.64)(320,22.88)(340,18.64)(360,22.03)(380,22.03)(400,19.49)(500,17.80)(600,16.10)(700,16.95)(800,21.19)(900,19.49)(1000,19.49)};

\nextgroupplot[title={ResNet50}]
\addplot+[mark=*, thick] coordinates {(0,1.69)(20,38.98)(40,33.90)(60,39.83)(80,39.83)(100,38.98)(120,40.68)(140,38.98)(160,43.22)(180,41.53)(200,38.14)(220,46.61)(240,43.22)(260,44.07)(280,42.37)(300,44.07)(320,38.98)(340,43.22)(360,42.37)(380,39.83)(400,44.07)(500,44.92)(600,40.68)(700,38.14)(800,43.22)(900,43.22)(1000,40.68)};
\addplot+[mark=square*, thick] coordinates {(0,1.69)(20,14.41)(40,18.64)(60,18.64)(80,21.19)(100,18.64)(120,21.19)(140,22.03)(160,22.88)(180,22.03)(200,20.34)(220,21.19)(240,22.88)(260,23.73)(280,24.58)(300,24.58)(320,25.42)(340,23.73)(360,22.88)(380,25.42)(400,23.73)(500,23.73)(600,22.88)(700,26.27)(800,22.88)(900,22.88)(1000,21.19)};
\addplot+[mark=triangle*, thick] coordinates {(0,2.54)(20,25.42)(40,36.44)(60,39.83)(80,41.53)(100,40.68)(120,39.83)(140,44.07)(160,42.37)(180,42.37)(200,49.15)(220,48.31)(240,50.85)(260,48.31)(280,47.46)(300,48.31)(320,46.61)(340,49.15)(360,49.15)(380,52.54)(400,50.00)(500,47.46)(600,52.54)(700,51.69)(800,48.31)(900,51.69)(1000,50.00)};
\addplot+[mark=diamond*, thick] coordinates {(0,1.69)(20,16.10)(40,20.34)(60,20.34)(80,24.58)(100,22.03)(120,22.88)(140,22.03)(160,23.73)(180,22.88)(200,22.03)(220,24.58)(240,23.73)(260,22.88)(280,22.88)(300,22.88)(320,21.19)(340,24.58)(360,22.88)(380,20.34)(400,21.19)(500,22.88)(600,22.03)(700,19.49)(800,24.58)(900,23.73)(1000,21.19)};

\nextgroupplot[title={VGG13}]
\addplot+[mark=*, thick] coordinates {(0,0.85)(20,21.19)(40,25.42)(60,24.58)(80,27.97)(100,22.03)(120,26.27)(140,24.58)(160,23.73)(180,21.19)(200,23.73)(220,26.27)(240,22.03)(260,25.42)(280,27.12)(300,23.73)(320,23.73)(340,23.73)(360,21.19)(380,26.27)(400,27.97)(500,27.12)(600,27.12)(700,27.97)(800,27.12)(900,27.12)(1000,28.81)};
\addplot+[mark=square*, thick] coordinates {(0,0.85)(20,13.56)(40,17.80)(60,16.10)(80,14.41)(100,15.25)(120,16.10)(140,16.10)(160,17.80)(180,15.25)(200,15.25)(220,14.41)(240,16.95)(260,16.95)(280,16.95)(300,17.80)(320,16.95)(340,15.25)(360,16.95)(380,17.80)(400,17.80)(500,16.10)(600,16.95)(700,17.80)(800,18.64)(900,15.25)(1000,16.95)};
\addplot+[mark=triangle*, thick] coordinates {(0,0.00)(20,25.42)(40,27.97)(60,27.97)(80,27.12)(100,27.97)(120,27.12)(140,29.66)(160,29.66)(180,27.97)(200,29.66)(220,27.97)(240,30.51)(260,28.81)(280,30.51)(300,30.51)(320,27.12)(340,30.51)(360,29.66)(380,31.36)(400,29.66)(500,27.12)(600,29.66)(700,30.51)(800,29.66)(900,28.81)(1000,31.36)};
\addplot+[mark=diamond*, thick] coordinates {(0,0.85)(20,16.10)(40,12.71)(60,15.25)(80,13.56)(100,11.86)(120,14.41)(140,13.56)(160,15.25)(180,14.41)(200,15.25)(220,14.41)(240,15.25)(260,15.25)(280,16.10)(300,16.10)(320,13.56)(340,15.25)(360,14.41)(380,14.41)(400,15.25)(500,15.25)(600,14.41)(700,14.41)(800,13.56)(900,15.25)(1000,13.56)};

\nextgroupplot[title={VGG16}]
\addplot+[mark=*, thick] coordinates {(0,1.69)(20,28.81)(40,32.20)(60,25.42)(80,33.05)(100,33.05)(120,33.05)(140,36.44)(160,34.75)(180,33.90)(200,33.05)(220,35.59)(240,37.29)(260,34.75)(280,23.73)(300,38.98)(320,31.36)(340,30.51)(360,36.44)(380,32.20)(400,31.36)(500,29.66)(600,34.75)(700,38.98)(800,32.20)(900,31.36)(1000,36.44)};
\addplot+[mark=square*, thick] coordinates {(0,1.69)(20,16.95)(40,14.41)(60,15.25)(80,15.25)(100,14.41)(120,13.56)(140,15.25)(160,16.10)(180,16.10)(200,16.10)(220,16.10)(240,15.25)(260,14.41)(280,15.25)(300,17.80)(320,15.25)(340,16.10)(360,16.95)(380,16.10)(400,15.25)(500,15.25)(600,15.25)(700,17.80)(800,16.95)(900,15.25)(1000,15.25)};
\addplot+[mark=triangle*, thick] coordinates {(0,1.69)(20,29.66)(40,32.20)(60,34.75)(80,33.05)(100,33.90)(120,35.59)(140,33.05)(160,36.44)(180,36.44)(200,34.75)(220,34.75)(240,34.75)(260,33.90)(280,37.29)(300,34.75)(320,35.59)(340,32.20)(360,36.44)(380,33.90)(400,35.59)(500,36.44)(600,39.83)(700,33.05)(800,35.59)(900,36.44)(1000,37.29)};
\addplot+[mark=diamond*, thick] coordinates {(0,1.69)(20,15.25)(40,15.25)(60,13.56)(80,16.10)(100,16.10)(120,15.25)(140,16.10)(160,14.41)(180,14.41)(200,15.25)(220,15.25)(240,13.56)(260,16.10)(280,14.41)(300,15.25)(320,15.25)(340,15.25)(360,16.10)(380,14.41)(400,16.95)(500,13.56)(600,15.25)(700,15.25)(800,15.25)(900,16.10)(1000,15.25)};

\end{groupplot}
\path (current bounding box.south) ++(0,-0.5cm) node[anchor=north] {
    \begin{tikzpicture}
        \begin{axis}[
            hide axis,
            xmin=0, xmax=1,
            ymin=0, ymax=1,
            legend style={
                at={(0.5,1)},
                anchor=south,
                legend columns=4,
                column sep=0.5cm,
                /tikz/every even column/.append style={column sep=0.5cm}
            }
        ]
        \addlegendimage{color=blue, mark=*, thick}
        \addlegendentry{Direction}
        \addlegendimage{color=red, mark=square*, thick}
        \addlegendentry{No Direction}
        \addlegendimage{color=brown!50!black, mark=triangle*, thick}
        \addlegendentry{Only Eyes}
        \addlegendimage{color=black, mark=diamond*, thick}
        \addlegendentry{Random}

        \end{axis}
    \end{tikzpicture}
};
\end{tikzpicture}
\caption{Accuracy vs. Epochs at learning rate $1 \times 10^{-5}$ for ResNet and VGG models under various annotation strategies.}
\end{figure}
\begin{figure}[H]
\centering
\begin{tikzpicture}
\begin{groupplot}[
    group style={
        group size=2 by 3,
        horizontal sep=1.2cm,
        vertical sep=2cm,
    },
    width=6.5cm,
    height=5.5cm,
    xlabel={Epochs},
    ylabel={Accuracy (\%)},
    grid=major,
    xtick={0, 200, 400, 600, 800, 1000},
    ymin=0, ymax=65,
    legend style={font=\small},
    legend cell align={left},
]

\nextgroupplot[title={ResNet18}]
\addplot+[mark=*, thick] coordinates {(0,3.39)(20,27.12)(40,31.36)(60,32.20)(80,34.75)(100,37.29)(120,31.36)(140,37.29)(160,33.05)(180,31.36)(200,38.14)(220,37.29)(240,38.14)(260,38.14)(280,35.59)(300,42.37)(320,34.75)(340,38.98)(360,38.98)(380,37.29)(400,41.53)(500,37.29)(600,37.29)(700,33.05)(800,37.29)(900,35.59)(1000,36.44)};
\addplot+[mark=square*, thick] coordinates {(0,3.39)(20,6.78)(40,12.71)(60,13.56)(80,16.10)(100,18.64)(120,15.25)(140,15.25)(160,16.95)(180,17.80)(200,17.80)(220,19.49)(240,18.64)(260,17.80)(280,16.95)(300,20.34)(320,17.80)(340,18.64)(360,17.80)(380,18.64)(400,17.80)(500,19.49)(600,18.64)(700,17.80)(800,18.64)(900,19.49)(1000,17.80)};
\addplot+[mark=triangle*, thick] coordinates {(0,2.54)(20,8.47)(40,20.34)(60,27.12)(80,35.59)(100,32.20)(120,33.90)(140,36.44)(160,37.29)(180,34.75)(200,41.53)(220,38.98)(240,42.37)(260,42.37)(280,38.98)(300,38.98)(320,41.53)(340,42.37)(360,40.68)(380,42.37)(400,43.22)(500,42.37)(600,48.31)(700,47.46)(800,44.92)(900,42.37)(1000,44.07)};
\addplot+[mark=diamond*, thick] coordinates {(0,3.39)(20,6.78)(40,12.71)(60,14.41)(80,15.25)(100,15.25)(120,17.80)(140,16.95)(160,17.80)(180,16.95)(200,16.95)(220,18.64)(240,18.64)(260,18.64)(280,16.95)(300,17.80)(320,19.49)(340,18.64)(360,16.10)(380,18.64)(400,21.19)(500,16.95)(600,19.49)(700,19.49)(800,18.64)(900,18.64)(1000,20.34)};

\nextgroupplot[title={ResNet34}]
\addplot+[mark=*, thick] coordinates {(0,0.85)(20,44.92)(40,45.76)(60,42.37)(80,41.53)(100,51.69)(120,46.61)(140,46.61)(160,44.92)(180,47.46)(200,44.07)(220,45.76)(240,44.07)(260,45.76)(280,45.76)(300,46.61)(320,45.76)(340,50.00)(360,44.07)(380,49.15)(400,48.31)(500,48.31)(600,44.92)(700,48.31)(800,43.22)(900,44.92)(1000,44.92)};
\addplot+[mark=square*, thick] coordinates {(0,0.85)(20,9.32)(40,12.71)(60,16.10)(80,16.10)(100,21.19)(120,17.80)(140,17.80)(160,17.80)(180,21.19)(200,22.03)(220,19.49)(240,20.34)(260,17.80)(280,21.19)(300,22.03)(320,20.34)(340,21.19)(360,21.19)(380,22.03)(400,22.03)(500,21.19)(600,21.19)(700,21.19)(800,22.88)(900,21.19)(1000,22.03)};
\addplot+[mark=triangle*, thick] coordinates {(0,0.85)(20,16.10)(40,32.20)(60,43.22)(80,43.22)(100,46.61)(120,49.15)(140,50.00)(160,49.15)(180,50.00)(200,49.15)(220,53.39)(240,50.85)(260,53.39)(280,50.00)(300,50.85)(320,50.85)(340,50.85)(360,52.54)(380,55.08)(400,50.85)(500,50.85)(600,55.93)(700,51.69)(800,54.24)(900,52.54)(1000,55.08)};
\addplot+[mark=diamond*, thick] coordinates {(0,1.69)(20,13.56)(40,14.41)(60,16.10)(80,19.49)(100,22.03)(120,22.88)(140,19.49)(160,18.64)(180,21.19)(200,17.80)(220,18.64)(240,22.03)(260,19.49)(280,20.34)(300,20.34)(320,18.64)(340,19.49)(360,18.64)(380,22.03)(400,22.03)(500,18.64)(600,19.49)(700,19.49)(800,22.03)(900,22.03)(1000,18.64)};

\nextgroupplot[title={ResNet50}]
\addplot+[mark=*, thick] coordinates {(0,1.69)(20,35.59)(40,36.44)(60,38.14)(80,41.53)(100,41.53)(120,39.83)(140,39.83)(160,44.07)(180,38.14)(200,44.92)(220,42.37)(240,38.14)(260,39.83)(280,42.37)(300,37.29)(320,45.76)(340,41.53)(360,41.53)(380,45.76)(400,42.37)(500,44.92)(600,41.53)(700,44.07)(800,43.22)(900,38.98)(1000,46.61)};
\addplot+[mark=square*, thick] coordinates {(0,1.69)(20,10.17)(40,14.41)(60,18.64)(80,20.34)(100,21.19)(120,18.64)(140,22.88)(160,21.19)(180,20.34)(200,21.19)(220,21.19)(240,23.73)(260,22.03)(280,23.73)(300,22.88)(320,21.19)(340,22.03)(360,26.27)(380,22.03)(400,22.88)(500,22.03)(600,22.88)(700,22.03)(800,24.58)(900,23.73)(1000,25.42)};
\addplot+[mark=triangle*, thick] coordinates {(0,2.54)(20,16.10)(40,22.03)(60,29.66)(80,35.59)(100,36.44)(120,40.68)(140,38.14)(160,35.59)(180,40.68)(200,44.07)(220,42.37)(240,44.07)(260,43.22)(280,41.53)(300,45.76)(320,45.76)(340,44.07)(360,46.61)(380,44.92)(400,42.37)(500,53.39)(600,50.85)(700,48.31)(800,51.69)(900,47.46)(1000,50.00)};
\addplot+[mark=diamond*, thick] coordinates {(0,1.69)(20,11.02)(40,16.10)(60,18.64)(80,16.95)(100,19.49)(120,20.34)(140,21.19)(160,22.88)(180,22.88)(200,23.73)(220,23.73)(240,21.19)(260,22.03)(280,22.88)(300,22.88)(320,22.03)(340,23.73)(360,23.73)(380,21.19)(400,21.19)(500,22.03)(600,24.58)(700,19.49)(800,22.88)(900,20.34)(1000,22.03)};

\nextgroupplot[title={VGG13}]
\addplot+[mark=*, thick] coordinates {(0,0.85)(20,21.19)(40,24.58)(60,27.12)(80,27.12)(100,23.73)(120,22.03)(140,25.42)(160,21.19)(180,24.58)(200,26.27)(220,27.12)(240,27.97)(260,27.12)(280,27.12)(300,27.97)(320,22.88)(340,25.42)(360,28.81)(380,24.58)(400,26.27)(500,27.97)(600,27.12)(700,27.97)(800,26.27)(900,27.97)(1000,25.42)};
\addplot+[mark=square*, thick] coordinates {(0,0.85)(20,14.41)(40,15.25)(60,14.41)(80,16.95)(100,17.80)(120,16.95)(140,16.95)(160,18.64)(180,16.10)(200,18.64)(220,16.10)(240,15.25)(260,16.10)(280,16.95)(300,16.95)(320,16.10)(340,16.10)(360,16.95)(380,16.10)(400,16.10)(500,18.64)(600,16.95)(700,16.95)(800,17.80)(900,17.80)(1000,16.95)};
\addplot+[mark=triangle*, thick] coordinates {(0,0.00)(20,17.80)(40,21.19)(60,26.27)(80,25.42)(100,26.27)(120,27.97)(140,27.97)(160,28.81)(180,26.27)(200,27.12)(220,27.97)(240,28.81)(260,28.81)(280,27.97)(300,27.97)(320,27.97)(340,28.81)(360,29.66)(380,27.97)(400,28.81)(500,29.66)(600,27.97)(700,27.97)(800,29.66)(900,31.36)(1000,29.66)};
\addplot+[mark=diamond*, thick] coordinates {(0,0.85)(20,16.10)(40,16.10)(60,12.71)(80,14.41)(100,15.25)(120,14.41)(140,13.56)(160,12.71)(180,13.56)(200,12.71)(220,14.41)(240,13.56)(260,12.71)(280,13.56)(300,14.41)(320,14.41)(340,14.41)(360,15.25)(380,16.95)(400,15.25)(500,16.10)(600,16.10)(700,15.25)(800,15.25)(900,15.25)(1000,15.25)};

\nextgroupplot[title={VGG16}]
\addplot+[mark=*, thick] coordinates {(0,1.69)(20,35.59)(40,29.66)(60,31.36)(80,30.51)(100,26.27)(120,31.36)(140,28.81)(160,30.51)(180,29.66)(200,29.66)(220,35.59)(240,37.29)(260,33.90)(280,32.20)(300,33.05)(320,33.05)(340,29.66)(360,31.36)(380,31.36)(400,28.81)(500,31.36)(600,35.59)(700,36.44)(800,31.36)(900,36.44)(1000,30.51)};
\addplot+[mark=square*, thick] coordinates {(0,1.69)(20,14.41)(40,16.10)(60,16.95)(80,14.41)(100,14.41)(120,15.25)(140,15.25)(160,15.25)(180,14.41)(200,14.41)(220,16.10)(240,15.25)(260,15.25)(280,16.95)(300,14.41)(320,16.10)(340,16.10)(360,16.95)(380,16.95)(400,14.41)(500,15.25)(600,16.10)(700,16.10)(800,15.25)(900,15.25)(1000,15.25)};
\addplot+[mark=triangle*, thick] coordinates {(0,1.69)(20,23.73)(40,30.51)(60,29.66)(80,31.36)(100,35.59)(120,33.90)(140,32.20)(160,35.59)(180,34.75)(200,33.90)(220,35.59)(240,33.90)(260,34.75)(280,37.29)(300,34.75)(320,36.44)(340,33.90)(360,36.44)(380,35.59)(400,34.75)(500,36.44)(600,36.44)(700,36.44)(800,37.29)(900,36.44)(1000,34.75)};
\addplot+[mark=diamond*, thick] coordinates {(0,1.69)(20,15.25)(40,16.95)(60,14.41)(80,15.25)(100,16.10)(120,15.25)(140,16.95)(160,15.25)(180,16.10)(200,13.56)(220,16.10)(240,15.25)(260,14.41)(280,16.10)(300,14.41)(320,15.25)(340,15.25)(360,15.25)(380,16.10)(400,15.25)(500,16.10)(600,15.25)(700,14.41)(800,15.25)(900,14.41)(1000,16.10)};

\end{groupplot}
\path (current bounding box.south) ++(0,-0.5cm) node[anchor=north] {
    \begin{tikzpicture}
        \begin{axis}[
            hide axis,
            xmin=0, xmax=1,
            ymin=0, ymax=1,
            legend style={
                at={(0.5,1)},
                anchor=south,
                legend columns=4,
                column sep=0.5cm,
                /tikz/every even column/.append style={column sep=0.5cm}
            }
        ]
        \addlegendimage{color=blue, mark=*, thick}
        \addlegendentry{Direction}
        \addlegendimage{color=red, mark=square*, thick}
        \addlegendentry{No Direction}
        \addlegendimage{color=brown!50!black, mark=triangle*, thick}
        \addlegendentry{Only Eyes}
        \addlegendimage{color=black, mark=diamond*, thick}
        \addlegendentry{Random}

        \end{axis}
    \end{tikzpicture}
};
\end{tikzpicture}
\caption{Accuracy vs. Epochs at learning rate $5 \times 10^{-6}$ for ResNet and VGG models under various annotation strategies.}
\end{figure}
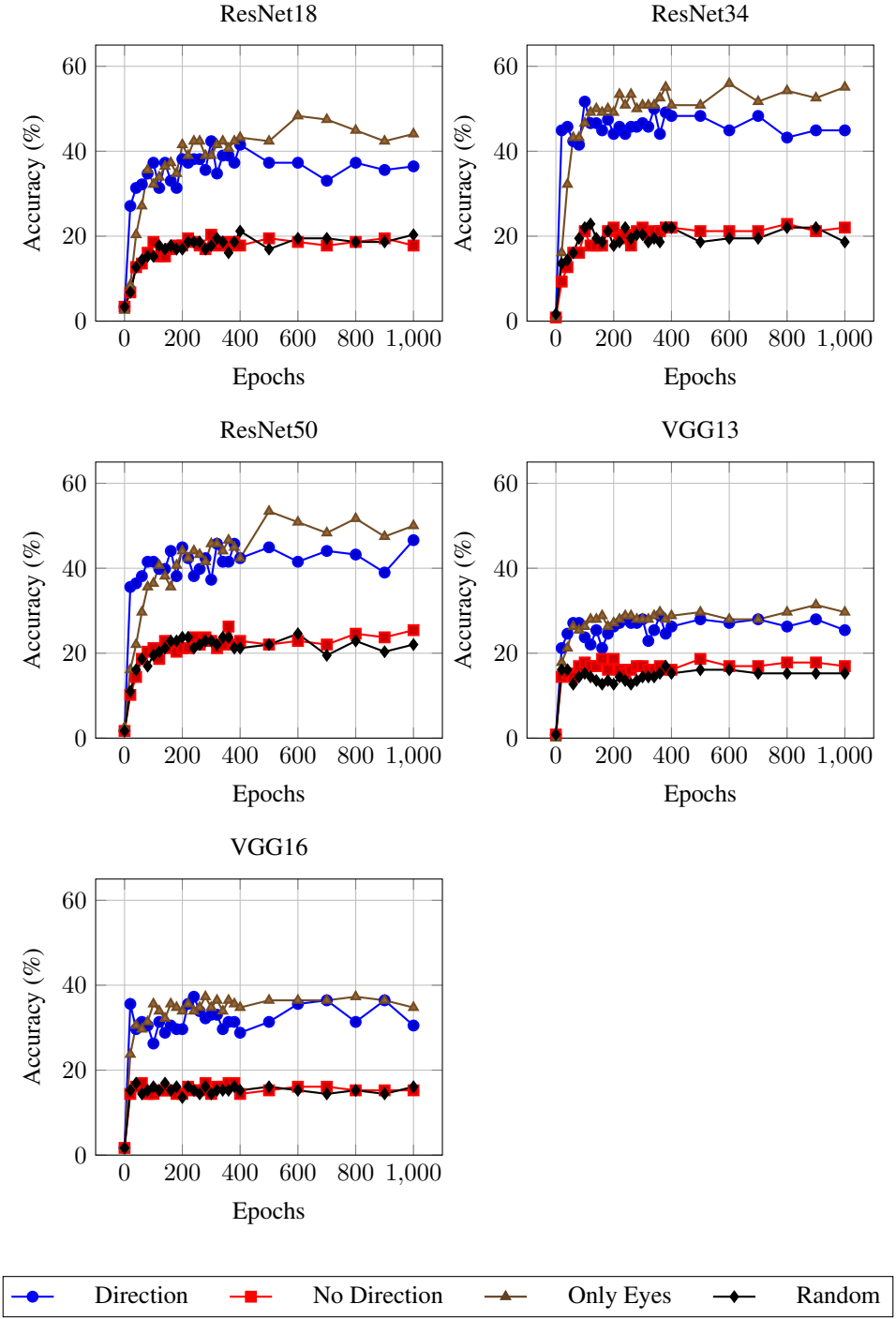

\begin{figure}[H]
    \centering
    \includegraphics[width=4cm]{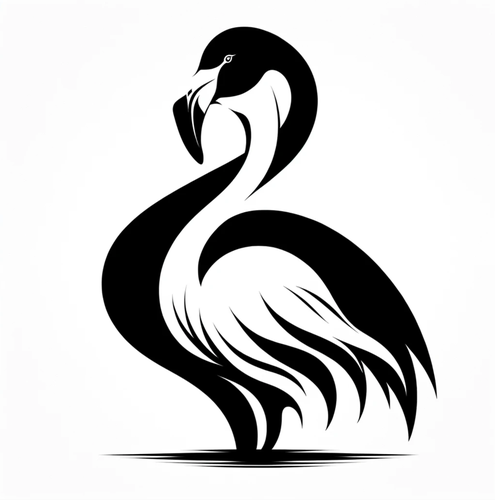}
    \caption{This is an example of a generated image where the gazes are completely shared and do not help in distinguishing which is the correct feature. Is there a more prominent answer on which one is seen with a higher likelihood, and what is it dependent on? The outer one? Do we prefer the color black? Is it the animal whose head \enquote{looks complete}? Does this change when we turn the angle of the picture, so another kind of \enquote{gaze direction}?}
    \label{fig:swan}
\end{figure}

Furthermore, we discussed that the explanations generated when including direction as a concept showed more useful features for recognition. Due to page restriction, we only included one example of this phenomenon in the main paper for the algorithm LIME \cite{lime}. However, we have other such examples for for example Gradcam \cite{selvaraju2017grad} and PipNet \cite{nauta2023pip}, which we will show and elaborate in the following: 
\begin{figure}[H]
    \centering
    \includegraphics[width=0.9\linewidth]{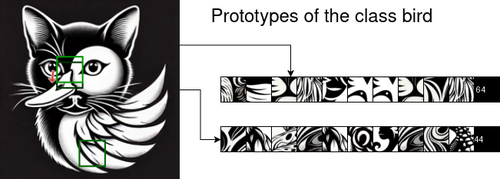}
    \caption{This image shows us the top prototypes marked by Pipnet for the class "bird". Again, Pipnet marks areas in the image that are important to the classification process. We can see that one of the boxes could either be the beak or the eye with the arrow, and one is focused on the feathers/wing patterns. Both are good features for the recognition of the animal. We also show some of the example patterns that Pipnet gives us for the feather pattern by which it classifies as bird. Even though this works well, this is yet another XAI method that focuses on areas instead of abstract concepts like viewing direction. }
    \label{fig:pipnet_arrowXAI}
\end{figure}
\begin{figure}[H]
    \centering
    \includegraphics[width=10cm]{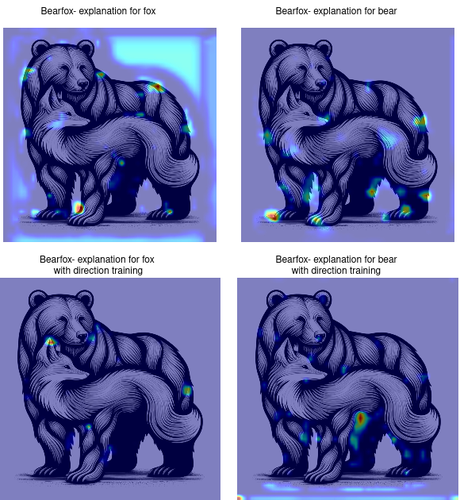}
    \caption{In this image, we can see with Gradcam on an example image how the learner was able to extract much more useful features for the model trained with the direction vector. The whiteness is removed, and for the fox explanations it focuses less on areas that are actually part of the bear. Regarding the bear explanation, less parts of the fox are highlighted. }
    \label{fig:bearfox}
\end{figure}
To go on, we noted in our limitations that we were able to generate some images that did not strictly fall into our scheme: These images included examples of animals where more than one animal is hidden in the body of an animal, which can be seen in Figure\ref{fig:turtle}: 
\begin{figure} [H]
    \centering
    \includegraphics[width=4cm]{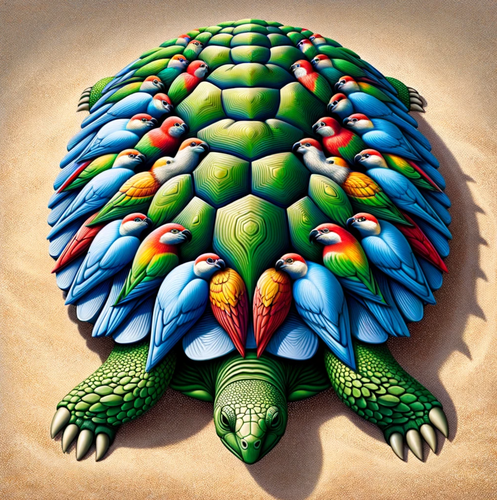}
    \caption{This shows us how easily optical illusions can be extended to more than just two animals within each other. We include several of these images generated accidentally when trying to generate as many different optical illusions as possible.}
    \label{fig:turtle}
\end{figure}
We must also consider what happens when the gaze is not the distinguishing feature, such as in Figure \ref{fig:swan}. Overall, this paper aims to broaden our perspective: What if highlighting pixels was the wrong approach for explainable AI in the image domain? What concepts should we really learn? Can we tackle more comprehensive learning strategies with the use of optical illusions and knowledge from psychological domains? These extra images will be included in the opensource dataset in a separate folder.

\end{document}